\documentclass[10pt,doublecolumn]{IEEEtran}
\usepackage{amsmath,amsthm,amsfonts,amssymb,bm,mathrsfs}
\usepackage{graphicx,subfigure}
\usepackage{cite}
\usepackage{algorithm}
\usepackage{algorithmic}
\usepackage{setspace}
\usepackage{multirow,booktabs,threeparttable,array}
\usepackage{tablefootnote}
\usepackage{color}
\usepackage{verbatim}
\usepackage{balance}
\graphicspath{{fig/}}

\theoremstyle{remark}

\theoremstyle{plain}

\allowdisplaybreaks[4]

\begin{document}
\title{The Gradient Convergence Bound of Federated Multi-Agent Reinforcement Learning with Efficient Communication}

\author{Xing Xu\textsuperscript{\rm 1}, Rongpeng Li\textsuperscript{\rm 1*}, Zhifeng Zhao\textsuperscript{\rm 2}, and Honggang Zhang\textsuperscript{\rm 1,\rm2}\\
	\textsuperscript{\rm 1}Zhejiang University, \textsuperscript{\rm 2}Zhejiang Lab\\
	\{hsuxing, lirongpeng, honggangzhang\}@zju.edu.cn, zhaozf@zhejianglab.com
}

\maketitle
\thispagestyle{empty}
\begin{abstract}
	The paper considers independent reinforcement learning (IRL) for multi-agent collaborative decision-making in the paradigm of federated learning (FL). However, FL generates excessive communication overheads between agents and a remote central server, especially when it involves a large number of agents or iterations. Besides, due to the heterogeneity of independent learning environments, multiple agents may undergo asynchronous Markov decision processes (MDPs), which will affect the training samples and the model's convergence performance. On top of the variation-aware periodic averaging (VPA) method and the policy-based deep reinforcement learning (DRL) algorithm (i.e., proximal policy optimization (PPO)), this paper proposes two advanced optimization schemes orienting to stochastic gradient descent (SGD): 1) A decay-based scheme gradually decays the weights of a model's local gradients with the progress of successive local updates, and 2) By representing the agents as a graph, a consensus-based scheme studies the impact of exchanging a model's local gradients among nearby agents from an algebraic connectivity perspective. This paper also provides novel convergence guarantees for both developed schemes, and demonstrates their superior effectiveness and efficiency in improving the system's utility value through theoretical analyses and simulation results.
\end{abstract}


\section{Introduction}
With the development of wireless communication and advanced machine learning technologies in the past few years, a large amount of data has been generated by smart devices and can enable a variety of multi-agent systems, such as smart road traffic control \cite{Ata2020Adaptive}, smart home energy management \cite{A2017A}, and the deployment of unmanned aerial vehicles (UAVs) \cite{Schwarzrock2018Solving,Parunak2004Digital,Cimino2016Combining}. Through deep reinforcement learning (DRL), an intelligent agent can gradually improve the performance of its parameterized policy via the trial-and-error interaction with the environment \cite{Volodymyr2015Human,John2015Trust,John2017Proximal,Tsallis2019Kyungjae}. However, directly applying DRL to multi-agent systems commonly faces several challenging problems, such as the non-stationary learning environment and the difficulty of reward assignment \cite{1996Reinforcement,Arulkumaran2017Deep}. As an alternative, independent reinforcement learning (IRL) is often employed in practical applications to alleviate the above-mentioned problems, where each agent undergoes an independent learning process with only self-related observations \cite{Tan1993Multi}. 

For each IRL agent, the training samples are obtained by going through a trajectory with the predefined terminal state or a certain number of Markov state transitions from Markov decision processes (MDPs). With the obtained samples, an agent can improve its performance by updating its policy's parameters along the gradient descent direction. In this paper, a policy-based DRL algorithm (i.e., proximal policy optimization (PPO) \cite{John2017Proximal}) is applied to calibrate the loss function, by repetitively calculating the gradients\cite{Sutton1998Reinforcement}. Generally speaking, the performance of DRL is closely related to the amount and variety of obtained samples, since the more fully explored state space leads to more accurate estimation of the cumulative reward signal \cite{1996Reinforcement}. Meanwhile, for a multi-agent system with naturally distributed IRL agents, the locally calculated policy gradients need to be shared through a coordination channel. Therefore, in order to effectively improve the performance of DRL, we adopt a federated multi-agent reinforcement learning (FMARL) framework, by combining the stochastic gradient descent (SGD) in DRL and federated learning (FL) \cite{Xing2019Stigmergic,H2020Optimizing,Cha2020Proxy}. In particular, a central server in FL is leveraged by iteratively aggregating the policy gradients from multiple IRL agents and in turn providing updated policy parameters. Therefore, FL can facilitate the implementation of this coordination channel and significantly contribute to enriching the sample information of each IRL agent indirectly. 

However, targeting this FMARL framework, there may be a large number of agents or policy iterations during the training phase. This naive implementation of FL may generate excessive communication overheads between agents and the central server. To alleviate this problem, periodic averaging has been naturally considered and popularly applied in many studies \cite{Chaudhari2017Parle,Smith2016CoCoA,Hao2018Parallel}, in which agents are allowed to perform several local updates within a period before transmitting their local gradients to the central server, so as to reduce the frequency of information exchange. However, an increase in the number of local updates would influence the convergence performance. Therefore, appropriate optimization methods should be developed to better balance the reduction of communication overheads and the improvement of the convergence performance. Besides, considering the heterogeneity of independent learning environments, multiple IRL agents may spend various amounts of time on state transition processes, and perform different numbers of local updates in the same period under the periodic averaging method, thus possibly affecting the convergence performance as well. Therefore, variations in the number of local updates under the periodic averaging method should be also considered.

Taking account of the model's error convergence bound with respect to the mainly required communication and computation overheads during the training phase, this paper proposes a system utility function-based metric to evaluate the effectiveness of different optimization methods for federated IRL (FIRL). Furthermore, in order to improve the system's utility value, this paper develops two new optimization methods (i.e., the decay-based and consensus-based methods) on top of the variation-aware periodic averaging (VPA) method. 
% Although similar theoretical analyses about the variation-aware periodic averaging method have been discussed in \cite{Hao2018Parallel} and \cite{Towards2020Kyungjae}, this paper begins with relevant theoretical results to facilitate the analysis of the developed methods. 
In particular, since the variance of model's gradients gradually increases along with the progress of local updating, the decay-based method utilizes a discrete periodic decay function to decrease the weights of a model's successive local gradients within one period, which also contrasts with the adoption of decaying learning rates for multiple epochs in \cite{Hao2018Parallel}. We concretely provide a practical implementation of this decay-based method with an exponential decay function and theoretically demonstrate its superior effectiveness. On the other hand, the consensus-based method introduces the consensus algorithm \cite{R2007Consensus} into both IRL and FL, where agents are allowed to exchange their local gradients with neighbors directly before performing local updates. Different from the spectral radius-based theoretical analyses in \cite{Multi-Stage2020Seyyedali}, this paper obtains the consensus-based method's error convergence bound dependent on the algebraic connectivity of the graph comprised by agents and their connections. We theoretically show that the consensus-based method can reduce the model's error convergence bound dramatically and speed up the learning process. Table \ref{tbr} summarizes the key differences with highly related works. Finally, through an MARL simulation scenario, we demonstrate the superiority of these two developed methods in terms of the model's convergence performance and system's utility value.

\begin{table*}[htpb]
	\footnotesize
	\centering
	\renewcommand\arraystretch{1.2}
	\caption{\label{tbr}Summary of the differences with highly related papers.}
	\begin{tabular}{m{0.45\linewidth}|m{0.45\linewidth}}
		\toprule[1.1pt]
		\multicolumn{2}{c}{\textbf{Articles combining FL and DRL}}                                                                                                                                                \\
		\hline
		\textbf{\cite{Su2022UAV}}: \textit{Decoupled} setting: hierarchical nested personalized FL for stratified UAVs swarms; DRL for swarm trajectory and learning duration design.                          & 
		\multirow{2}{\linewidth}{\textbf{Ours}: Integrated setting: \textit{Direct enhancement} of decentralized \textit{multi-agent} DRL by FL.}                                                                 \\
		\cline{1-1}
		\textbf{\cite{Yang2023Optimizing}}: \textit{Single-agent} DRL-based \textit{device selection} for asynchronous FL.                                                                                     & 
		~                                                                                                                                                                                                         \\
		\hline
		\textbf{\cite{Xie2023FedKL}}: Targeting at the \textit{data heterogeneity} issue of multi-agent DRL by imposing KL divergence-based penalty term.                                                      & 
		\textbf{Ours}: Targeting at the \textit{learning inefficiency} issue of multi-agent DRL by decay and consensus-based FL methods.                                                                          \\
		\midrule
		\multicolumn{2}{c}{\textbf{Decay-based Method}}                                                                                                                                                           \\
		\midrule		
		\textbf{\cite{Hao2018Parallel}}: Decaying the learning rate of \emph{different epochs}.                                                                                                                & 
		\textbf{Ours}: Decaying the weights of different mini-bathes within \emph{one period}.                                                                                                                    \\
		\hline 
		\textbf{\cite{Towards2020Kyungjae}}: Assuming an \emph{off-line} available set of data points and assigning \emph{aggregation weights according to their sizes}. No consideration of the
		resource cost.                                                                                                                                                                                         & 
		\textbf{Ours}: \emph{On-line} DRL tasks and assigning time-decreasing \emph{aggregation
		weights regardless of data sizes}.                                                                                                                                                                        \\
		\midrule
		\multicolumn{2}{c}{\textbf{Consensus-based Method}}                                                                                                                                                       \\
		\midrule
		\textbf{\cite{Jiang2017Collaborative,H2020Decentralized,Xiangru2018Asynchronous}}: \emph{Disabled} central-server's functionality and only consensus-based aggregation of different agents' gradients. & 
		\textbf{Ours}: \emph{Hierarchical} configuration with both the agent-server interaction
		and agents' mutual interaction.                                                                                                                                                                           \\
		\hline
		\textbf{\cite{Multi-Stage2020Seyyedali}}: Convergence result with respect to the \emph{spectral radius} of the 
		network topology.                                                                                                                                                                                      & 
		\textbf{Ours}: Convergence result with respect to the \emph{algebraic connectivity}
		of the network topology.                                                                                                                                                                                  \\
		\bottomrule[1.1pt]
	\end{tabular}
\end{table*}

In summary, the main contributions of this paper are summarized as follows.
\begin{itemize}
	\item We propose an on-line FMARL framework, by improving the policy performance of distributed IRL agents through FL. Considering the possible excessive communication overheads and heterogeneous independent learning environments, we take the VPA method as the basis of systematic analysis.
	\item In order to measure and optimize the effectiveness of policy-based DRL (i.e., PPO) through general SGD \cite{Sutton1998Reinforcement}, we put forward a system's utility function, which quantifies the convergence bound of the model's error reduction per unit of resource cost during the training phase, so as to reasonably evaluate the effectiveness of different optimization methods.
	\item On top of the systematic analysis in this paper, we propose some novel implementation means to realize the decay-based and consensus-based methods for FIRL, so as to more efficiently aggregate the gradients from multiple agents within each period. Specifically, an exponentially decaying function is applied to decrease the weights of successive local gradients, while the gradients from multiple agents are merged together in a hierarchical manner. We demonstrate the superiority of both methods through theoretical analyses and numerical simulation results.
\end{itemize}

The remainder of this paper is mainly organized as follows. In Section \ref{sec:relatedworks}, we explain the related works and clarify the novelty of our work. In Section \ref{sec:preliminaries}, we present preliminaries of the periodic averaging method. In Section \ref{sec:model}, we introduce the system model and formulate the optimization problem. In Section \ref{sec:method}, we consider the possible heterogeneity of FMARL and analyze the VPA method with different numbers of updates each period. In Section \ref{sec:bound}, we describe two optimization methods (i.e., the decay-based and consensus-based methods) and give their error convergence bounds. In Section \ref{sec:result}, we introduce the MARL simulation scenario and present the corresponding results of the developed methods. In Section \ref{sec:conclusion}, we conclude this paper with a summary. 

\section{Related Works}
\label{sec:relatedworks}
In recent years, distributed IRL \cite{Tan1993Multi} has been frequently studied and applied in many practical applications\cite{Littman1994Markov,K2007Distributed,Busoniu2008A}. However, due to the heterogeneity of independent learning environments, the performance of IRL agents may vary significantly, even though agents are deployed in the same global environment and face exactly the same facilities. Several studies \cite{Mnih2016Asynchronous,John2017Proximal,Sartoretti2018Distributed} have shown that experiences from homogeneous independent learning agents in the same multi-agent system can be collected and sampled together to obtain a shared model efficiently. Besides, the recent studies \cite{Xing2019Stigmergic,H2020Optimizing,Cha2020Proxy} have combined distributed IRL with FL to improve the involved agents' capability and collaboration efficiency. The combining of FL and DRL remains a hot research topic \cite{Yang2023Optimizing,Su2022UAV,Xie2023FedKL}. For example, \cite{Su2022UAV} adopts a decoupled setting to combine DRL and FL. Wherein, \cite{Su2022UAV} uses a hierarchical nested personalized FL for stratified UAVs swarms, while leverages DRL for swarm trajectory and learning duration design. \cite{Yang2023Optimizing} focuses on a digital-twin empowered industrial Internet-of-things (IoTs) scenario, and presents a single-agent DRL-based device selection for asynchronous FL. Contrary to these papers, our work is primarily oriented at the direct enhancement of decentralized multi-agent DRL by FL. Furthermore, \cite{Xie2023FedKL} targets at the data heterogeneity issue of multi-agent DRL by imposing KL divergence-based penalty term. Instead, our work aims to solve learning inefficiency issue of multi-agent DRL by decay and consensus-based FL methods. Besides, we further quantify the effect of key parameters in FL on the convergence performance.

An on-line federated transfer reinforcement learning framework is introduced in \cite{Liang2019Federated} based on the deep deterministic policy gradient (DDPG) \cite{Silver2014Deterministic} algorithm for autonomous driving. However, this framework lacks the theoretical analysis for the performance of FL in DRL, while the interplay between deterministic actor network and the Q-function typically makes DDPG extremely difficult to stabilize and brittle to hyper-parameter settings \cite{Tuomas2019Soft}. Meanwhile, a federated DRL-based cooperative edge caching (FADE) framework is proposed in \cite{Wang2020Federated} to optimize the edge caching schemes in IoTs services. However, this framework ignores that the frequent information exchange between base stations (BSs) and UEs may generate excessive communication overheads. Moreover, FADE utilizes a value-based DRL method (i.e., deep Q-learning network (DQN) \cite{Hado2015Deep}) and theoretically analyzes the model's convergence in SGD based on some general classification loss functions. Nevertheless, the value-based DRL methods exhibit some instability issues in high-dimensional scenarios. Besides, the effect of deploying SGD in improving DRL under these functions is hard to characterize. Instead, this paper utilizes the periodic averaging method to further reduce the excessive communication overheads, and deploys the policy-based DRL method (i.e., PPO algorithm \cite{John2017Proximal}) to effectively improve DRL by SGD \cite{Sutton1998Reinforcement}.

FL is a parallelly distributed machine learning paradigm, aiming to train specific model through the samples distributed across different agents. Due to the protection requirement for data privacy and the restriction from communication bandwidth or delay \cite{Rui2020CPFed}, FL allows distributed agents to calculate their models' gradients locally, and then forward these gradients towards a central server to centrally update the model's parameters. However, the naive FL may generate excessive communication overheads between distributed agents and the central server, especially when there are a large number of agents or iterations \cite{Peter2019Advances}. Hence, the periodic averaging method has been proposed to make agents locally perform several updates to the model within a period before transmitting local gradients to the server \cite{Jakub2016Federated,Wang2018Cooperative,S2018When,Haddadpour2019Local}. Moreover, several quantization or sparsification methods have been also proposed to implement lossy compression for the model's local gradients that need to be transmitted \cite{Wu2018Error,Wang2018ATOMO,CommunicationShi2020,Intermittent2020Wang}. Furthermore, FL with flexible device participation has been considered in \cite{Towards2020Kyungjae}, where devices (i.e., agents) are assumed to have different sample sizes during the training phase. However, \cite{Towards2020Kyungjae} is not designed for on-line training tasks. Instead, \cite{Towards2020Kyungjae} assumes devices already have available data points and could assign aggregation weights according to their sizes. On the contrary, in this paper, the learning paradigm is oriented to on-line DRL tasks, and the discrepancy of agents in terms of sample-collecting efficiency and computing capability is additionally considered. Specifically, we reflect this discrepancy on the numbers of agents' local updates, and formulate our problem on the VPA method.

The theoretical analyses about the error convergence bound under the periodic averaging method have been provided in \cite{Hao2018Parallel}. Moreover, to reduce this error convergence bound, \cite{Hao2018Parallel} and \cite{Towards2020Kyungjae} have proposed to decrease the learning rate over epochs and optimize the aggregation weights for agents with different sample sizes, respectively. Differently, the decay-based method in this paper aims to decrease the weights of successive gradients over local updates within one period. Note that each epoch includes several periods. Moreover, we additionally provide a practical implementation of the decay-based method through an exponential function, in which the distribution of  agents' local updates is further considered. On the other hand, the consensus algorithm \cite{R2007Consensus} has been applied in many decentralized averaging methods \cite{Jiang2017Collaborative,H2020Decentralized,Xiangru2018Asynchronous}, which normally disable the central server's functionality but allow agents to directly exchange and average their parameters with neighbors. However, decentralized averaging methods will slow the convergence rate as the size of agents' network increases. A consensus-based FL algorithm has been proposed in \cite{Multi-Stage2020Seyyedali} to obtain the training model in large-scale device-to-device (D2D)-enabled fog networks, and an upper bound of convergence with respect to the spectral radius (i.e., the largest eigenvalue of the built induced consensus matrix) of the network topology has been also derived. On the contrary, we derive the error convergence bound from the algebraic connectivity (i.e., the second smallest eigenvalue of the Laplace matrix). Together with our previous work \cite{Xu2022Trustable}, this bound additionally reflects the effect of local interaction metric (i.e., step size $\epsilon$ in the classical consensus algorithm \cite{R2007Consensus}), which can be properly adjusted to guarantee the policy improvement of DRL. Finally, both the decay-based and consensus-based methods are oriented to on-line MARL tasks, where theoretical analyses about the error convergence bound are based on general assumptions for the VPA method.

\begin{table}
	\footnotesize
	\centering
	\renewcommand\arraystretch{1.0}
	\caption{Main Notations Used in This Paper.}
	\label{tb1}
	\begin{tabular}{ll}
		\toprule[1.1pt]
		Notation              & Definition                                         \\
		\toprule[1.1pt]
		$\pi$                 & Stochastic policy                                  \\
		$s,\ s_t$             & Local state                                        \\
		$\mathcal{S}$         & Local state space                                  \\
		$a,\ a_t$             & Individual action                                  \\
		$\mathcal{A}$         & Individual action space                            \\
		$K$                   & Total number of iterations                         \\
		$\eta$                & Proper learning rate                               \\
		$\mathcal{L}$         & Loss function                                      \\
		$\phi_t$              & Markov state transition                            \\
		$r$                   & Individual reward                                  \\
		$\mathcal{R}$         & Reward function                                    \\
		$\xi_k$               & Mini-batch                                         \\
		$N$                   & Total number of agents                             \\
		$m$                   & Maximal number of selected agents                  \\
		$\tau$                & Number of local updates within a period            \\
		$T$                   & Maximal length of an epoch                         \\
		$U$                   & Number of epochs                                   \\
		$P$                   & Maximal length of a step                           \\
		$C_1$                 & Communication overheads                            \\
		$C_2$                 & Computation overheads                              \\
		$F$                   & Empirical risk function                            \\
		$\psi_0$              & Resource cost under general settings               \\
		$\psi_0^{(\text{C})}$ & Resource cost under the consensus-based VPA method \\
		$\psi_1$              & Error bound of the learning model                  \\
		$\psi_1^{(\text{P})}$ & Error bound under the periodic averaging method    \\
		$\psi_1^{(\text{V})}$ & Error bound under the VPA method                   \\
		$\psi_1^{(\text{D})}$ & Error bound under the decay-based VPA method       \\
		$\psi_1^{(\text{C})}$ & Error bound under the consensus-based method       \\
		$\psi_2$              & Expected error of the initial model                \\
		$L$                   & Lipschitz constant                                 \\
		$\beta,\ \sigma^2$    & Non-negative constants                             \\
		$\nu$                 & Mean value of the number of local updates          \\
		$w^2$                 & Variance value of the number of local updates      \\
		$\lambda$             & Decay constant                                     \\
		$E$                   & Total number of local interactions                 \\
		$G$                   & Topology of agents' network                        \\
		$\Omega_i$            & Set of neighbors                                   \\
		$\epsilon$            & Local interaction step size                        \\
		$\mathbf{La}$         & Laplace matrix                                     \\
		$\mu$                 & Eigenvalue of Laplace matrix                       \\
		$W_1$                 & Communication overheads in local interaction       \\
		$W_2$                 & Computation overheads in local interaction         \\
		\bottomrule[1.1pt]
	\end{tabular}
\end{table}

\section{Preliminaries}
\label{sec:preliminaries}
Main notations used in this paper are listed in Table \ref{tb1}. We assume that each IRL agent maintains a DRL model with parameterized policy $\pi(s, a;\theta)$, where $s \in \mathcal{S}$ is sampled from the local state space, $a \in \mathcal{A}$ is selected from the individual action space, $\pi : \mathcal{S} \times \mathcal{A} \to [0,1]$ is a stochastic policy, and $\mathbf{\theta} \in \mathbb{R}^{d}$, where $d$ denotes the dimension of parameters. According to stochastic policy gradient methods for the policy optimization process of DRL \cite{John2015Trust}, the learning model's parameters $\mathbf{\theta}$ can be updated by
\begin{equation}
	\label{eq:theta_update}
	\mathbf{\theta}_{k+1} = \mathbf{\theta}_{k} - \eta\nabla_{\mathbf{\theta}_{k}}\mathbf{\mathcal{L}},
\end{equation}
where $k$ denotes the index of policy iteration and $\eta$ is the learning rate with a reasonable step size. According to the PPO algorithm \cite{John2017Proximal}, $\mathbf{\mathcal{L}}(\theta)$ is defined as $\mathcal{L}(\theta) = \mathbb{E}_t \left[ -J_t^{\textrm{Clip}}(\theta) + c_1 J_t^{V}(\theta) - c_2 S(s_t;\theta) \right]$, where $c_1$, $c_2$ are weighting factors and $S$ is an entropy bonus. $J_t^{V}(\theta)=\left(V_{\theta}(s_t) - V^{\textrm{Target}}(s_t)\right)^2$ is a squared-error loss, where $V^{\textrm{Target}}$ denotes the target state value function. $J_t^{\textrm{Clip}}(\theta)$ denotes a clipped objective and is defined as $J_t^{\textrm{Clip}}(\theta) =  \textrm{min}\left(\frac{\pi_{\theta}(s_t, a_t)}{\pi_{\theta_{\textrm{old}}}(s_t, a_t)}A_t, \textrm{clip}\left(\frac{\pi_{\theta}(s_t, a_t)}{\pi_{\theta_{\textrm{old}}}(s_t, a_t)}, 1-\zeta, 1+\zeta\right)A_t\right)$, where $\zeta$ denotes the clipping parameter. $A_t=-V(s_t)+r_t+\gamma r_{t+1}+ \cdots  +\gamma^{T-t}V(s_T)$ denotes the accumulated advantage across multiple steps, where $\gamma$ is a discount factor. Here, $\mathbf{\mathcal{L}}$ denotes the loss function to be minimized, and facilitates the calculation for the gradients of objective functions in DRL. Commonly, SGD is used to optimize the loss function and \cite{Gradient2016Lee} proves SGD could converge for any twice-differentiable loss function, regardless of its convexity. Moreover, our following derivations are based on common assumptions (e.g., $L$-smooth bounded objective functions). Therefore, the results in this paper can be easily extended to typical settings.

In addition, to stand in consistent with DRL \cite{Volodymyr2015Human}, we define the sample used in the policy iteration as
\begin{equation}
	\phi_{t} :=\ \langle s_t,\ a_t,\ r_t,\ s_{t+1} \rangle,
	\label{sample}
\end{equation}
where $t$ denotes the time-stamp within an epoch, while an epoch has a predefined terminal state $s_{\textrm{terminal}}$ or a certain number of sequential Markov state transitions $\phi_{t}$. $r_t$ is calculated by a reward function $\mathcal{R} : \mathcal{S} \to \mathbb{R}$. In order to speed up the training process and reduce the variance of gradients resulting from a single sample, a certain number of samples are picked out at each iteration to comprise a mini-batch. Therefore, the practical gradients used for training are written as
\begin{equation}
	g(\theta_k;\xi_k) = \frac{1}{|\xi_k|} \sum_{\phi_{t} \in \xi_k} \nabla \mathbf{\mathcal{L}}(\theta_k;\phi_{t}),
\end{equation}
where $\xi_k$ denotes the mini-batch at iteration $k$, and $|\xi_k|$ represents its size. 

On the other hand, considering the multi-agent parallel training process in FL, we can obtain
\begin{equation}
	\label{eq:multi-agent-fl-parallel}
	\mathbf{\theta}_{k+1} = \mathbf{\theta}_{k} - \eta \frac{1}{m} \sum_{i=1}^{m} g(\theta_k;\xi_k^{(i)}),
\end{equation}
where $m$ denotes the maximal number of agents that transmit the local gradients to the central server at iteration $k$, and $\xi_k^{(i)}$ is the mini-batch from agent $i$. Furthermore, when the periodic averaging method is applied to aggregate local updates from agents, \eqref{eq:multi-agent-fl-parallel} can be updated as
\begin{equation}
	\mathbf{\theta}_{k+1}^{(i)} = 
	\begin{cases}
		{\frac{1}{m} \sum_{i=1}^{m} \left[\mathbf{\theta}_{k}^{(i)} - \eta g(\theta_k^{(i)};\xi_k^{(i)})\right]}, & \text{if\ $k\ \mathrm{mod}\ \tau = 0$}; \vspace{2ex} \\
		{\mathbf{\theta}_{k}^{(i)} - \eta g(\theta_k^{(i)};\xi_k^{(i)})},                                         & \text{otherwise},   
	\end{cases}
	\label{eq:multi-agent-fl-parallel_update}
\end{equation}
where $\tau$ represents the number of local updates in a period. For simplicity, we will use the notation $g(\theta_k^{(i)})$ to represent $g(\theta_k^{(i)};\xi_k^{(i)})$ in the rest of this paper. 

\section{System Model and Problem Formulation}
\label{sec:model}
\begin{figure}
	\centering
	\includegraphics[width=0.48\textwidth]{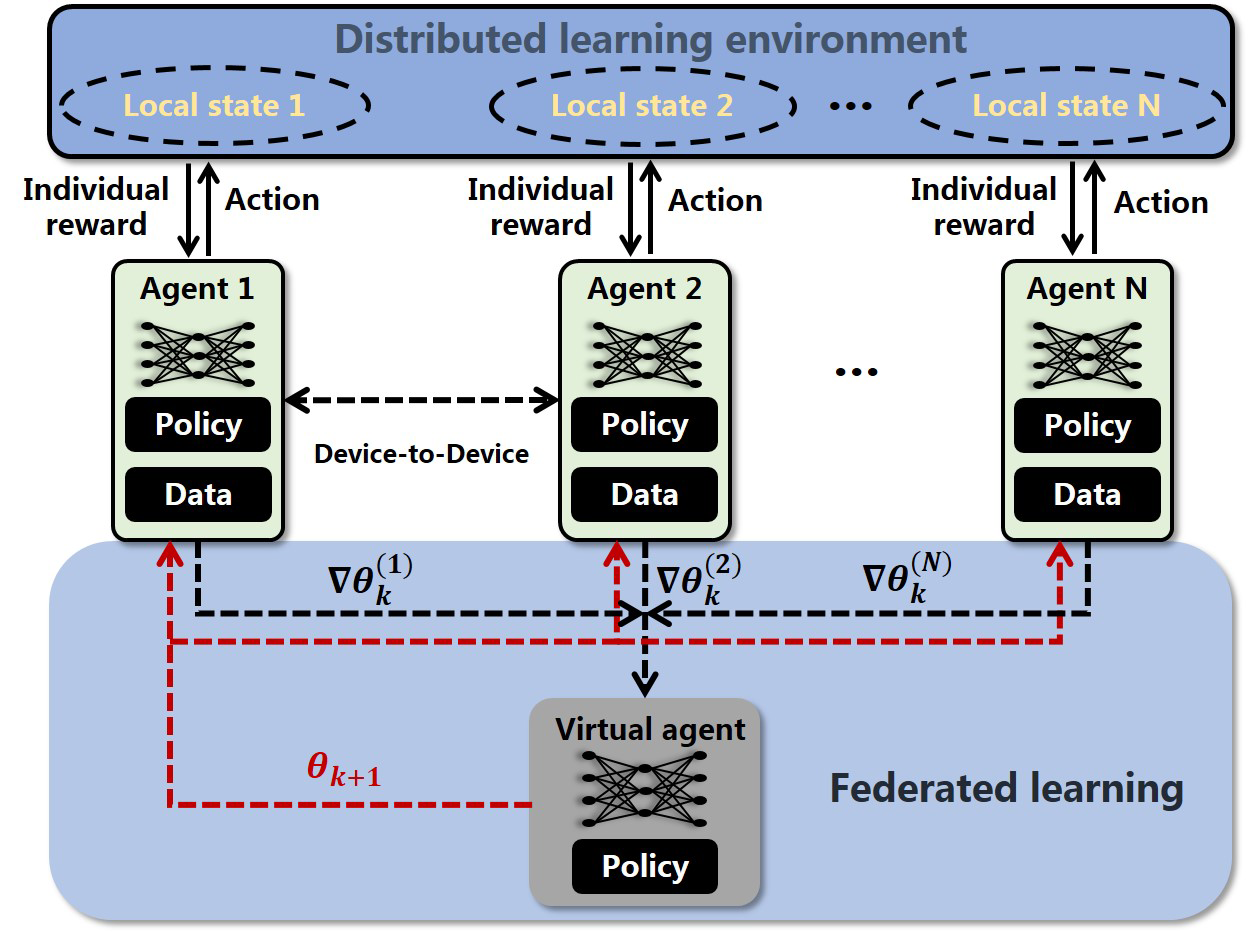}
	\caption{Framework of FMARL.}
	\label{fig1}
\end{figure}

Fig. \ref{fig1} illustrates how FMARL introduces FL into MARL. As depicted in Fig. \ref{fig1}, each IRL agent is designed to learn independently through FL in the training phase, and operate autonomously in the execution phase via DRL. Within this multi-agent distributed learning scenario, the role of a conventional central server is played by a virtual agent, which can be deployed at a remote cloud center or some capable local agent. Similar to decentralized POMDP \cite{Deep2017Shayegan}, we suppose that each agent can only observe a local state to characterize part of the global environment, and an individual reward will be returned to the agent immediately after an action being performed. Meanwhile, to facilitate the multi-agent collaboration, we allow agents to exchange their local gradients with nearby collaborators through the D2D communication whenever needed.

\begin{figure*}  
	\begin{minipage}{0.4\linewidth}
		\centerline{\includegraphics[width=1\linewidth]{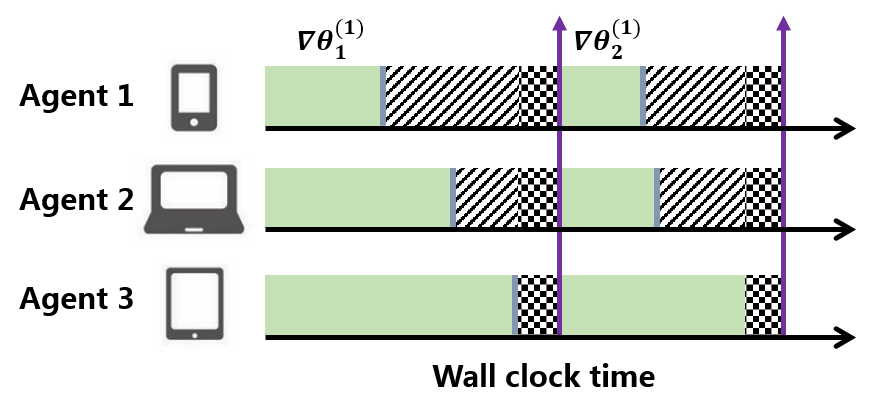}}
		\centerline{\small{(a)}}
	\end{minipage}
	\hfill
	\begin{minipage}{.4\linewidth}
		\centerline{\includegraphics[width=1\linewidth]{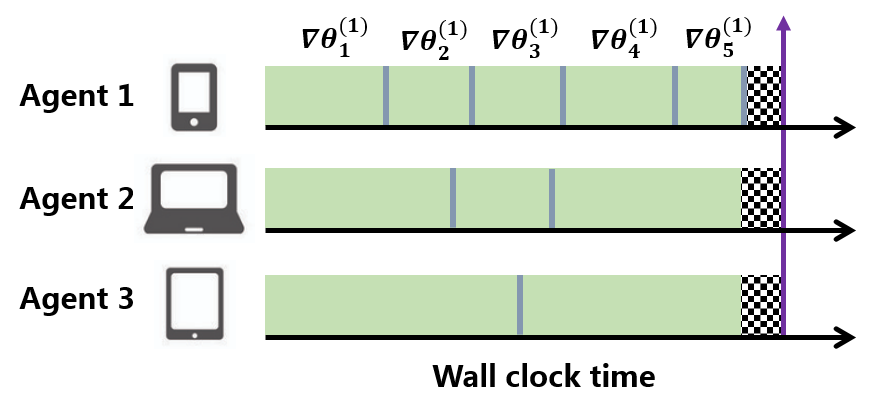}}
		\centerline{\small{(b)}}
	\end{minipage}
	\vfill
	\begin{minipage}{0.4\linewidth}
		\centerline{\includegraphics[width=1\linewidth]{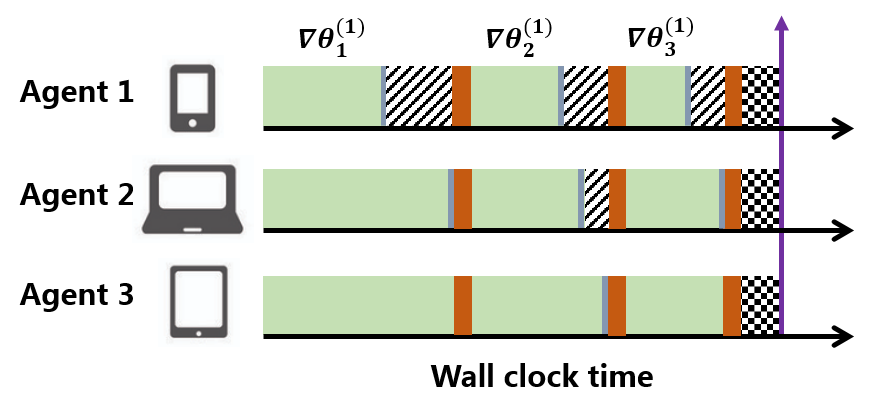}}
		\centerline{\small{(c)}}
	\end{minipage}
	\hfill
	\begin{minipage}{.4\linewidth}
		\centerline{\includegraphics[width=0.7\linewidth]{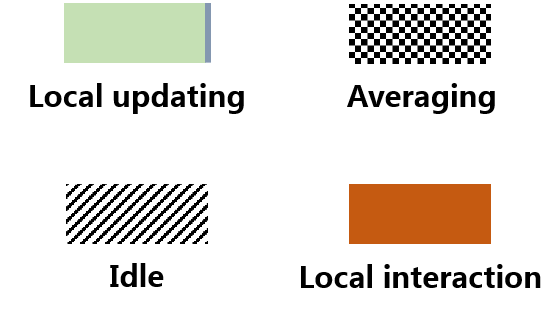}}
	\end{minipage}
	\caption{Intuitive schematic of (a) FL without periodic averaging, (b) the periodic averaging method with $\tau = 5$, and (c) the consensus-based periodic averaging method with $\tau = 3$.}
	\label{fig2}
\end{figure*}

\subsection{Local Updates}
Without loss of generality, we assume that there are totally $N$ agents involved in the multi-agent distributed learning scenario. Meanwhile, the maximal length of an epoch and the total number of epochs for training are denoted as $T$ and $U$, respectively. To facilitate the convergence rate of DRL \cite{Tesauro1995Temporal}, an epoch is further uniformly divided into several steps, each of which generally contains a predefined number of sequential Markov state transitions (i.e., samples in (\ref{sample})). In particular, each step includes $P$ transitions as a mini-batch and a local update could be obtained correspondingly. Furthermore, considering that different agents may spend various amounts of time on state transition processes, the wall clock time spent by agent $i$ to finish a step is denoted by a random variable $x_i$ for $i=1,2,...,N$, where $x_i \in \mathbb{R^+}$. Meanwhile, these random variables satisfy $\mathbb{E}[x_1] \le \mathbb{E}[x_2] \le \cdots \le \mathbb{E}[x_\mathrm{N}]$. Since the local update only happens after finishing the corresponding step, the number of local updates in a period (i.e., the interval between two successive periodic averaging) for each agent is determined by
\begin{equation}
	\tau_i := \left\lfloor \tau \frac{\mathbb{E}[x_{1}]}{\mathbb{E}[x_{i}]} \right\rfloor,
	\label{eqt}
\end{equation}
where $\tau$ indicates the number of local updates that agent $i=1$ can finish on average in a period and should be determined beforehand. $\lfloor \cdot \rfloor$ is the round down operation and $\tau_i \in \mathbb{N^+}$. Here, the duration of a period is dynamically determined through the time spent by agent $i=1$ to finish $\tau$ local updates. Therefore, for agent $i=1$, an epoch approximately consists of $\frac{T}{P\tau}$ periods. The reason for this design is that agent $i=1$ can experience the most transitions from the learning environment, and thus has the greatest weight on the training model. Moreover, in (\ref{eqt}), the computation time for any local update is ignored, as it is commonly much less than the time required to finish a certain number of sequential Markov state transitions in DRL. Besides, in order to reduce the training time, only the agents already completing at least one local update at the end of a period need to transmit their local gradients to the virtual agent. Consistent with the notations in \eqref{eq:multi-agent-fl-parallel}, we set the maximal number of these involved agents as $m$, where $m \le N$. In the rest of this paper, we will focus most of our attention on these $m$ agents. In Figs. \ref{fig2}(a) and (b), we provide an intuitive schematic of agents performing local updates without and with the periodic averaging method, respectively. It can be observed that the implementation of the periodic averaging method makes the numbers of local updates from different agents unsynchronized. Specifically, in \eqref{eq:multi-agent-fl-parallel} and (\ref{eq:multi-agent-fl-parallel_update}), the agent $i$ with a rather small $\tau_{i}$ may satisfy $\xi_k^{(i)}= \varnothing$ and $g(\theta_{k}^{(i)})=\mathbf{0}$ at some iteration $k$.

\subsection{Resource Cost}
Considering the resources spent by agents during the training phase of FMARL, we basically suppose that the main communication overheads required for an agent to transmit its local gradients to the virtual agent are $C_1$, while the main computation overheads to perform a local update are $C_2$. Then, similar to the definition of resource cost in \cite{Rui2020Differentially}, the system's resource cost shall reflect the effect of key parameters (e.g., $\tau$) in optimization methods on the resource cost, and can be intuitively formulated as
\begin{equation}
	\psi_0 = \sum_{i=1}^{m} \left(\frac{C_1TU}{\tau P} + \frac{C_2 \tau_i TU}{\tau P}\right).
	\label{psi0}
\end{equation}
It can be observed from (\ref{psi0}) that a larger averaging period (i.e., $\tau \neq 1$) can reduce the communication overheads (i.e., the first item in (\ref{psi0})) required for the involved agents by $\tau$ times, while a decrease in the number of local updates (i.e., $\tau_i \le \tau$) can reduce the computation overheads of each agent. However, (\ref{psi0}) intentionally neglects the impact of $\tau$ on the performance. Therefore, in order to evaluate the model's performance improvement per unit of resource cost, we shall introduce some preliminary results on the stochastic policy gradient methods as well.

\subsection{Convergence Bound of Stochastic Policy Gradient Methods}
For any epoch, the optimization objective of policy iteration can be expressed as
\begin{equation}
	\pi_{\theta} = \mathop{\arg\min}_{\theta \in \mathbb{R}^d} \mathbb{E}_t\left[\mathbf{\mathcal{L}}(\pi_{\theta};\phi_{t})\right].
	\label{obj}
\end{equation}
According to \eqref{eq:theta_update} and the definition of $\mathbf{\mathcal{L}}$, we accomplish the optimization objective in (\ref{obj}) by performing SGD for the policy parameters $\theta$. In the following discussions, we will replace the notation $\pi_{\theta}$ with $\theta$ for simplicity of representations. Furthermore, we define an empirical risk function as
\begin{equation}
	F(\theta) := \frac{1}{\left| \xi \right|} \sum_{\phi \in \xi} \mathbf{\mathcal{L}}(\theta;\phi).
\end{equation}
In practical applications, since the empirical risk function $F(\theta)$ may be non-convex, the learning model's convergence may fall into a local minimum or saddle point. In our framework, similar to \cite{Lian2015Asynchronous,Wang2018Cooperative,Bottou2018OptimizationMF}, the expected gradient norm is used as an indicator of the model's convergence to guarantee that it falls into a stationary point, that is
\begin{equation}
	\label{eq:psi_1_definition}
	\mathbb{E}\left[\frac{1}{K} \sum_{k=0}^{K-1} \left\|\nabla F(\bar{\theta}_k)\right\|^2\right] \le \psi_1,
\end{equation}
where $K$ denotes the expected total number of policy iterations, and $K = UT/P$. $\|\cdot\|$ denotes the $\ell_2$ vector norm. $\bar{\theta}_k$ represents the average parameters of all agents at iteration $k$, which is also regarded as the final result at the end of each iteration. Notably, $\psi_1$ denotes the targeted error convergence bound, towards reducing which we can optimize the means to update the parameters $\bar{\theta}_k$ (or $\theta_{k}^{(i)}$, for $i \in \{1,2,3,...,m\}$). Moreover, $k \mod \tau = 0$, and the model with $\bar{\theta}_k$ is maintained and updated by the virtual agent as
\begin{equation}
	\bar{\theta}_k = \bar{\theta}_{t_0} - \eta \frac{1}{m} \sum_{i=1}^{m} \sum_{y=t_0}^{k - 1} \mathrm{I}(\tau_i > y - t_0) \ g(\theta_{y}^{(i)}),
	\label{virtual}
\end{equation}
where $t_0 = z\tau$, $z \in \mathbb{N}$, and $t_0$ denotes the index of the iteration, at which the virtual agent performs the latest periodic averaging before iteration $k$. $\mathrm{I}(\cdot)$ denotes the indicator function. Here, each agent transmits the accumulated gradients within a period (i.e., $\sum_{y=t_0}^{k - 1} \mathrm{I}(\tau_i > y - t_0) g(\theta_{y}^{(i)})$) to the virtual agent, so as to update the parameters $\bar{\theta}_k$ in a centralized manner. Similarly, we define $\bar{\theta}_0$ as the model's initial parameters for all agents and obtain
\begin{equation}
	\psi_2 := \mathbb{E}\left[\left\|\nabla F(\bar{\theta}_0)\right\|^2\right],
\end{equation}
where $\psi_2$ denotes the expected gradient norm of the initial model, which represents the initial convergence error and is gradually decreased by SGD. 

\subsection{Optimization Objective of FMARL}
Inspired by the optimization objective in \cite{Rui2020Differentially}, which utilizes a predefined threshold as an upper bound to restrict the value of resource cost $\psi_0$, we formulate the optimization objective of FMARL as
\begin{align}
	\mathop{max}\limits_{\bar{\theta}_k}\ \mathcal{U} & := \frac{\psi_2 - \psi_1}{\psi_0}, \label{maxi}                                                           \\
	\textrm{s.t.}\  \psi_1                            & \ge \mathbb{E}\left[\frac{1}{K} \sum_{k=0}^{K-1} \left\|\nabla F(\bar{\theta}_k)\right\|^2\right], \notag \\
	\psi_2                                            & = \mathbb{E}\left[\left\|\nabla F(\bar{\theta}_0)\right\|^2\right], \notag                                \\
	\psi_0                                            & = \sum_{i=1}^{m} \left(\frac{C_1TU}{\tau P} + \frac{C_2 \tau_i TU}{\tau P}\right),\notag
\end{align}
where $\mathcal{U}$ denotes the system's utility function, and indicates the lower bound (rather than the exact value) of error reduction per unit of resource cost. Given the model's initial parameters $\bar{\theta}_0$, $\mathcal{U}$ is jointly determined by the targeted error convergence bound $\psi_1$ and the resource cost $\psi_0$. In order to increase $\mathcal{U}$, we can equivalently minimize $\psi_1$ or $\psi_0$. Hence, we propose to primarily optimize the update means of $\bar{\theta}_k$ in two different optimization methods, aiming to further reduce the error convergence bound $\psi_1$, but 1) keep the resource cost $\psi_0$ unchanged (i.e., the decay-based method), or 2) make the resource cost $\psi_0$ not increase too much (i.e., the consensus-based method).

\section{Variation-aware Periodic Averaging Method}
\label{sec:method}
In this section, we derive the model's error convergence bound under the VPA method and highlight some useful observations. We consider the heterogeneity of independent learning environments faced by IRL agents from the perspective of the number of local updates $\tau$. Note that one mini-batch consisting of $P$ Markov transitions corresponds to one local update. Consistent with the theoretical analyses about the periodic averaging method in \cite{Hao2018Parallel} and the situation where agents have different sample sizes in \cite{Towards2020Kyungjae}, we begin with relevant theoretical results under some common assumptions.

\subsection{General Assumptions}
In this paper, the learning model's error convergence bound is inferred under the following general assumptions, which are similar to those presented in previous studies for distributed SGD \cite{Wang2018Cooperative,Bottou2018OptimizationMF}.

\noindent
\rm\textbf{Assumption 1 (\underline{A1})}
\begin{itemize}
	\setstretch{1.4}
	\item[1.] (Smoothness): $\left\|\nabla F(\theta) - \nabla F(\theta^{'})\right\| \le L\left\|\theta - \theta^{'}\right\|$;
	\item[2.] (Lower bounded): $F(\theta) \ge F_{\mathrm{inf}}$;
	\item[3.] (Unbiased gradients): $\mathbb{E}_{\theta|\xi}\left[g(\theta)\right] = \nabla F(\theta)$;
	\item[4.] (Bounded variance): $\mathbb{E}_{\theta|\xi}\left[\left\|g(\theta) - \nabla F(\theta)\right\|^{2}\right] \le \beta\left\|\nabla F(\theta)\right\|^{2} + \sigma^{2}$,
\end{itemize}
where $L$ represents the Lipschitz constant, which implies that the empirical risk function $F$ is $L$-smooth \cite{Zhou2018Duality}. $F_{\mathrm{inf}}$ denotes the lower bound of $F$, and we suppose that it can be reached when the total number of iterations $K$ is large enough, which satisfies $F(\bar{\theta}_K) = F_{\mathrm{inf}}$. In addition, $\beta$ and $\sigma^2$ are both non-negative constants and inversely proportional to the size of mini-batch \cite{Wang2018Cooperative}. Condition 3 and 4 in \underline{A1} on the bias and variance of the mini-batch gradients are general for SGD methods \cite{Rui2020CPFed}. Specifically, the variance of the mini-batch gradients is bounded by Condition 4 through the value that fluctuates with the exact gradients rather than through a constant as in previous studies \cite{Hao2018Parallel,Rui2020CPFed}, which thus sets up a looser restriction. Note that \underline{A1} is the fundamental assumption that will be applied to other analyses in this paper.

\begin{algorithm}[t]
	\small
	\caption{\textbf{The training algorithm of FMARL.}}
	\begin{algorithmic}[1]
		\REQUIRE the model's initial parameters $\bar{\theta}_0$;
		\ENSURE the model's final average parameters $\bar{\theta}_k$;
		\STATE \textbf{Initialize} entire environment, learning rate $\eta$, loss function $\mathcal{L}$, reward function $\mathcal{R}$, maximal length of an epoch $T$, total number of epochs for training $U$, maximal size of a mini-batch $P$, number of local updates $\tau$ for agent $i=1$, number of agents that need to transmit the model's local gradients $m$, and iteration index $k$;
		\FOR {epoch $u=1,2,3,...,U$}
		\FOR {transition $t=0,1,2,...,T-1$}
		\FOR {agent $i=1,2,3,...,m$}
		\STATE Calculate the input vector $s_t$ according to the received local state from the environment;
		\STATE Select an action $a_t$ according to $\theta_k^{(i)}(s_t,\ a_t)$;
		\STATE Perform the selected action and receive the next state $s_{t+1}$ from the environment;
		\STATE Calculate $r_t$ according to the reward function $\mathcal{R}$;
		\STATE Store this transition as $\phi_t^{(i)}$;
		\IF {$t+1$ mod $P$ == 0 \textbf{or} $t$ == $T-1$}
		\STATE Form the mini-batch $\xi_k^{(i)}$ by the stored transitions $\phi_t^{(i)}$;
		\STATE Calculate the mini-batch gradients by \\ $g(\theta_k^{(i)}) = \frac{1}{|\xi_k^{(i)}|} \sum_{\phi_{t}^{(i)} \in \xi_k^{(i)}} \nabla \mathbf{\mathcal{L}}(\theta_k^{(i)};\phi_{t}^{(i)})$;
		\STATE Perform the local update by \\ $\mathbf{\theta}_{k+1}^{(i)} = \mathbf{\theta}_{k}^{(i)} - \eta g(\theta_k^{(i)})$;
		\STATE Clear the stored transitions $\phi_t^{(i)}$;
		\STATE Store the mini-batch gradients $g(\theta_k^{(i)})$;
		\STATE $k \gets k+1$;
		\ENDIF
		\IF {$k$ mod $\tau == 0$ $\textbf{or}$ reach the end of a period}
		\STATE Transmit all the accumulated gradients $g(\theta_y^{(i)})$ to the virtual agent and receive the model's average parameters by \\
		$\bar{\theta}_k = \bar{\theta}_{t_0} - \eta \frac{1}{m} \sum_{i=1}^{m} \sum_{y=t_0}^{k - 1} \mathrm{I}(\tau_i > y - t_0)\ g(\theta_{y}^{(i)})$;
		\STATE $\theta_k^{(i)} \gets \bar{\theta}_k$;
		\STATE Clear the stored transitions $\phi_t^{(i)}$;
		\STATE Clear the accumulated gradients $g(\theta_y^{(i)})$;
		\ENDIF
		\ENDFOR
		\ENDFOR
		\ENDFOR 
		\STATE \textbf{Return} the model's average parameters $\bar{\theta}_k$;
	\end{algorithmic}
\end{algorithm}

\subsection{Convergence Bound of FMARL with Same $\tau$}
During the training phase of DRL, it is common to keep the learning rate $\eta$ as a proper constant and decay it only when the performance saturates. Similarly, in the following discussions, we will investigate the model's error convergence bound under a fixed learning rate. We first discuss the situation where there is no difference in the number of local updates between the agents involved (i.e., $\forall i \in \{1,2,3,...,m\},\ \tau_i = \tau$). The training algorithm of FMARL under the periodic averaging method is illustrated in Algorithm 1. By performing Algorithm 1, we suppose that the training samples obtained by agents and the learning model satisfy \underline{A1}, and the total number of iterations $K$ is large enough and divisible by $\tau$. If the learning rate $\eta$ satisfies 
\begin{equation}
	\eta L (\frac{\beta}{m} + 1) - 1 + 2 \eta^{2} L^{2} \tau \beta + \eta^{2} L^{2} \tau (\tau + 1) \le 0,
	\label{t1}
\end{equation}
then the expected gradient norm after $K$ iterations is bounded by
\begin{equation}
	\begin{aligned}
		\mathbb{E}\left[\frac{1}{K} \sum_{k=0}^{K-1} \left\|\nabla F(\bar{\theta}_k)\right\|^{2}\right]  
		 & \le  \underbrace{\frac{2\left[F(\bar{\theta}_0) - F_{\mathrm{inf}}\right]}{\eta K}  + \frac{\eta L \sigma^{2}}{m}} \\ &\underbrace{+\ \eta^{2} L^{2} \sigma^{2} (\tau + 1)}_{\psi_1^{(\text{P})}}.
		\label{lb1}
	\end{aligned}
\end{equation}
(\ref{t1}) and (\ref{lb1}) indicate a practical approach for \eqref{eq:psi_1_definition} from the periodic averaging method. For the sake of distinction, we will use the notation $\psi_1^{(\text{P})}$ to represent the bound under the periodic averaging method (i.e., relative to $\psi_1$). The corresponding proofs about (\ref{t1}) and (\ref{lb1}) are provided in the Appendix B. 

\noindent
\emph{Remarks:} It can be observed that (\ref{t1}) indicates an upper bound for the learning rate $\eta$. In particular, we can get $\eta \in (0, \frac{1}{L(\tau + 1)}]$ when $\beta$ in (\ref{t1}) is close to $0$, which is practical when $\sigma^2$ is large \cite{Hao2018Parallel,Rui2020CPFed}. Since both $L$ and $\tau$ are positive constants, general settings of the learning rate $\eta$ in DRL (e.g., $1.0 \times 10^{-4}$) could meet the requirement, and the performance impact of $\eta$ will be further discussed in Section \ref{sec:result}. Besides, $\psi_1^{(\text{P})}$ shows that the model's error convergence bound is jointly influenced by several key parameters. 
Specifically, the first term of $\psi_1^{(\text{P})}$ satisfies $\frac{2\left[F(\bar{\theta}_0) - F_{\mathrm{inf}}\right]}{\eta K} \ge \frac{1}{K} \sum_{k=0}^{K-1} \left[ 2\langle \nabla F(\bar{\theta}_k), \mathcal{G}_k \rangle - \eta L \left\| \mathcal{G}_k \right\|^2 \right]$, where $\mathcal{G}_k = \frac{1}{m} \sum_{i=1}^{m} g(\theta_{k}^{(i)})$, and is affected by the calculated gradients $g(\theta_{k}^{(i)})$ as well as the Lipschitz constant $L$, which is related to the smooth properties of the empirical risk function $F$. Since $g(\theta_{k}^{(i)}) \rightarrow 0$ when $k \rightarrow K$, the first term of $\psi_1^{(\text{P})}$ finally approaches $0$.
In addition, the second term of $\psi_1^{(\text{P})}$ implies that an increase in the number of participating agents $m$ can reduce $\psi_1^{(\text{P})}$, but it comes at the expense of more resource cost and may reduce the system's utility value (see (\ref{psi0}) and (\ref{maxi})). Furthermore, we can also conclude that although the periodic averaging method can reduce the communication overheads by many times, the increase of $\tau$ would enlarge the error bound and affect the convergence rate \cite{Peter2019Advances}. Note that the result in (\ref{lb1}) is similar to those presented in existing works \cite{Rui2020CPFed,Hao2018Parallel,Wang2018Cooperative}, but we still provide it here as the preliminary conclusion, so as to facilitate subsequent theoretical analyses.

\subsection{Convergence Bound of FMARL with Different $\tau$}
Considering that different agents may spend various amounts of time finishing a step for each local update, we further derive the model's error convergence bound under the VPA method, where the variation comes from that the local updates from different agents are asynchronous, as illustrated in Fig. \ref{fig2}(b). Specifically, we make the following assumptions for the agents involved.

\noindent
\rm\textbf{Assumption 2 (\underline{A2})}
\begin{itemize}
	\setstretch{1.4}
	\item[1.] $\tau_i \in \{1,2,3,...,\tau\}$, for $i = 1,2,3,...,m$;
	\item[2.] $\tau_i \ge \tau_{i+1}$, for $i = 1,2,3,...,m-1$;
	\item[3.] $\sum_{i=1}^{m} \mathrm{I}(\tau_i = \tau) \ge 1$;
	\item[4.] $\frac{1}{m} \sum_{i=1}^{m} \tau_i = \bar{\tau}_{i} \stackrel{K \to \infty}{\longrightarrow} \nu$;
	\item[5.] $\frac{1}{m} \sum_{i=1}^{m} (\tau_{i} - \bar{\tau}_{i})^{2} \stackrel{K \to \infty}{\longrightarrow} \omega^{2}$.
\end{itemize}

\noindent
Note that Condition 1 and 2 in \underline{A2} on local updates are consistent with the previous definition of $\tau_{i}$ in (\ref{eqt}) within the scope of $m$ participating agents. Moreover, Condition 3 in \underline{A2} also conforms to the previous definition for the duration of a period, which is dynamically determined by the wall clock time required for some agent able to complete $\tau$ local updates. In addition, by Condition 4 and 5 in \underline{A2}, we suppose that the numbers of local updates from $m$ participating agents have the mean and variance value (i.e., $\nu$ and $\omega^2$) when the total number of iterations $K$ is large enough, so as to facilitate the convergence analysis.

Under the VPA method, the update rule of $\bar{\theta}_{k}$ at the virtual agent is determined by (\ref{virtual}). Similarly, the update rule of $\theta_{k}^{(i)}$ within a period at each agent is determined by
\begin{equation}
	\theta_{k}^{(i)} = \bar{\theta}_{t_0} - \eta \sum_{y=t_0}^{k - 1} \mathrm{I}(\tau_{i} > y - t_0)\ g(\theta_{y}^{(i)}).
\end{equation}
The training algorithm of FMARL under the VPA method can be also illustrated by Algorithm 1. By performing Algorithm 1, we suppose that the  learning model and the involved agents satisfy \underline{A1} and \underline{A2}, while the total number of iterations $K$ is large enough and divisible by $\tau$. If the learning rate $\eta$ satisfies (\ref{t1}), then the expected gradient norm after $K$ iterations is bounded by
\begin{align}
	\mathbb{E}\left[\frac{1}{K} \sum_{k=0}^{K-1} \left\|\nabla F(\bar{\theta}_k)\right\|^{2} \right] 
	\le  \underbrace{\frac{2[F(\bar{\theta}_0) - F_{\mathrm{inf}}]}{\eta K}  +  \frac{\eta L \sigma^{2}}{m}} \notag \\
	\underbrace{+\ \frac{\eta^{2} L^{2} \sigma^{2}}{\tau} \left[-\nu^{2} + (2\tau + 1)\nu - \omega^{2}\right]}_{\psi_1^{(\text{V})}}. \label{lb2}
\end{align}
The corresponding proofs are provided in the Appendix C. 

\noindent
\emph{Remarks:} Compared with the error convergence bound $\psi_1^{(\text{P})}$ in (\ref{lb1}), we can observe that the third term of $\psi_1^{(\text{V})}$ in (\ref{lb2}) reveals more details. Specifically, with the value of $\tau$ unchanged, the part in square brackets can be regarded as a quadratic function of the mean value $\nu$. Furthermore, we can find that the maximal value of this quadratic function would be reached at $\nu = \tau + 1/2$. Since \underline{A2} ensures $1 < \nu \le \tau$, we can conclude that the error convergence bound will increase monotonically as the mean value $\nu$ goes up, which is similar to the effect brought by the increase of $\tau$. We can also observe that an increase in the variance value $\omega^{2}$ can reduce the error convergence bound $\psi_1^{(\text{V})}$. Besides, if $\nu = \tau$ and $\omega = 0$, the VPA method will reduce to the classical periodic averaging method. Compared with the convergence bound obtained in \cite{Towards2020Kyungjae}, the conclusion in (\ref{lb2}) further considers the effect of the mean and variance value of $\tau$ on convergence performance. Next, we will take this VPA method as the basis of subsequent optimization methods.

\section{Convergence Bound of SGD Considering Communication Efficiency}
%In Section VI-A and VI-B, through the discussions for key parameters, we can finally propose two new optimization methods to improve the system's utility value. 
\label{sec:bound}
\subsection{Decay-based Method}
For a single agent, since there may be a deviation between the distribution of obtained samples and that of real inputs, the direction of SGD can thus contain error. Through averaging the model's local gradients in FL, this error can be reduced by boosting many gradient descent directions from different agents. However, the frequency of this averaging process is greatly reduced in the periodic averaging method, and the error of SGD will be superposed continuously with the progress of local updates, resulting in the variance of subsequent local gradients gradually increasing. To solve this problem, we take advantage of a decay function to gradually decrease the weights of successive local gradients within each period, which satisfies the following assumptions. 

\noindent
\rm\textbf{Assumption 3 (\underline{A3})}
\begin{itemize}
	\setstretch{1.4}
	\item[1.] $D(y) = D(y + \tau)$, and $y \in \mathbb{N}$;
	\item[2.] $1 = D(y = t_0) \ge D(y = t_0 + 1) \ge D(y = t_0 + 2) \ge \\\cdots \ge D(y = t_0 + \tau - 1) \ge 0$.
\end{itemize}
It can be observed that the decay function $D(y)$ is defined as a discrete periodic function, and is monotonically decreasing over a period of length $\tau$. Under the decay-based method, $D(y)$ is used to decay the weight of mini-batch gradients when updating the model's parameters. Therefore, the update rules of $\theta_{k}^{(i)}$ and $\bar{\theta}_k$ over a period of length $\tau$ are expressed as
\begin{equation}
	\theta_{k}^{(i)} = \bar{\theta}_{t_0} - \eta \sum_{y=t_0}^{k - 1} \mathrm{I}(\tau_{i} > y - t_0)\ D(y)\ g(\theta_{y}^{(i)}),
	\label{decayagent}
\end{equation}
\begin{equation}
	\bar{\theta}_k = \bar{\theta}_{t_0} - \eta \frac{1}{m} \sum_{i=1}^{m} \sum_{y=t_0}^{k - 1} \mathrm{I}(\tau_i > y - t_0)\ D(y)\ g(\theta_{y}^{(i)}).
	\label{decayvirtual}
\end{equation}
The training algorithm of FMARL under the decay-based method is almost the same as that illustrated in Algorithm 1, except that a common decay function needs to be provided for each agent in advance and the update rules of the model's parameters should follow (\ref{decayagent}) and (\ref{decayvirtual}). % For the sake of distinction, we will use the notation $\psi_1^{(\text{D})}$ to represent the model's error convergence bound under the decay-based method (i.e., relative to $\psi_1$). 
Then, we can draw the following theorem.

\noindent 
\rm\textbf{Theorem 1 (\underline{T1})} \textit{Suppose the number of local updates for agent $i$ is $\tau_i$, and the model's training process follows Algorithm 1, where agents update their parameters according to (\ref{decayagent}) and (\ref{decayvirtual}). Under  \underline{A1}, \underline{A2}, and \underline{A3}, if the total number of iterations $K$ is large enough and divisible by $\tau$, and the learning rate $\eta$ satisfies (\ref{t1}), then the upper bound of the expected gradient norm after $K$ iterations satisfies}
\begin{equation}
	\psi_1^{(\text{D})} \le \psi_1^{(\text{V})}.
	\label{t3}
\end{equation}
where $\psi_1^{(\text{D})}$ indicates the bound for the decay-based VPA method. The corresponding proofs are provided in the Appendix D.

\noindent
\emph{Remarks:} The gap between $\psi_1^{(\text{D})}$ and $\psi_1^{(\text{V})}$ depends on the applied decay function. On the other hand, considering the resource cost under the decay-based method, we can find that the proposed method can reduce the error convergence bound while maintaining the resource cost (i.e., $\psi_0$) unchanged, thus improving the system's utility value.

To further characterize the proposed decay-based method, we provide an example to highlight its advantages. In particular, we concretely define $D(y)$ as an exponential function, which is expressed by
\begin{equation}
	D(y) := \lambda^{\frac{y}{2}},
	\label{decay}
\end{equation}
where $\lambda \in (0,1]$ is a decay constant. To facilitate the convergence analysis, we also assume that the numbers of local updates from $m$ participating agents are uniformly distributed within the domain (i.e., $\mathrm{Pr}(\tau_{i} = \tau_{0}) = 1/\tau$, for $ \tau_{0} \in \{1,2,3,...,\tau\}$). This will be practical when $m$ is large and local agents are very diversified. Therefore, according to Condition 4 and 5 in \underline{A2}, we can easily get $\nu = (1+\tau)/2$ and $\omega^{2} = (\tau - 1)^{2}/12$. We further can get the following corollary.

\noindent
\rm\textbf{Corollary 1 (\underline{C1})} \textit{Under the conditions in \underline{T1}, suppose the numbers of agents' local updates are uniformly distributed within the domain, while the decay function is defined as (\ref{decay}). If the learning rate $\eta$ satisfies (\ref{t1}), then the expected gradient norm after $K$ iterations is bounded by\footnote{For sake of simplicity of representation, we slightly abuse the notation of $\psi_1^{(\text{D})}$ for both a general decay-based method and a specific exponential function-driven one.}}
\begin{align}
	\label{lb4}
	\mathbb{E}\left[\frac{1}{K} \sum_{k=0}^{K-1} \left\|\nabla F(\bar{\theta}_k)\right\|^{2}\right] \le \underbrace{\frac{2[F(\bar{\theta}_0) - F_{\mathrm{inf}}]}{\eta K}  +  \frac{\eta L \sigma^{2}}{m}} \notag \\
	\underbrace{+\ \frac{2 \eta^{2} L^{2} \sigma^{2}}{\tau} \left[\frac{\tau}{1-\lambda} - \frac{2\lambda}{(1-\lambda)^{2}} + \frac{\lambda(\lambda + 1)(1-\lambda^{\tau})}{\tau(1-\lambda)^{3}}\right]}_{\psi_1^{(\text{D})}}.
\end{align}
\noindent
The corresponding proofs about (22) are provided in the Appendix E.

\noindent
\emph{Remarks:} It can be observed that with the value of $\tau$ unchanged, the part in square brackets of the third term of $\psi_1^{(\text{D})}$ in (\ref{lb4}) can be regarded as a function of $\lambda$. Through calculating its first-order derivative, we can figure out that this function is monotonically increasing as the value of $\lambda$ increases. Moreover, the bound $\psi_1^{(\text{D})}$ obtained in \underline{C1} will be equal to $\psi_1^{(\text{P})}$ in (\ref{lb1}) when $\lambda \rightarrow 0$ and thus $\tau \rightarrow 1$. And it will also approach the bound obtained in (\ref{lb2}) when $\lambda \rightarrow 1$ and thus $D(y) \rightarrow 1$. By adjusting $\lambda$ in this decay-based method, we can indirectly control the maximal number of samples that agents can employ in a period for training. In practical applications, $\lambda$ can be set to a value that is slightly less than $1$, such as $\lambda = 0.98$.

\subsection{Consensus-based Method}
To take full advantage of the multi-agent collective collaboration, we further propose to employ the consensus algorithm \cite{R2007Consensus} to improve the local update of each agent through the D2D communication among nearby collaborators. Since the original objective of the consensus algorithm is to make all the distributed nodes in an ad-hoc network reach a consensus, it can be used to reduce the variance of the mini-batch gradients from a cluster of agents. Under the consensus-based method, we will use the notation $g(\theta_{k}^{(i)}, e)$ to distinguish different local interactions of agents towards optimizing $g(\theta_{k}^{(i)})$, where $e$ denotes the index of local interactions, and $g(\theta_{k}^{(i)}, 0) = g(\theta_{k}^{(i)})$. In addition, to enable all the participating agents to reach a consensus successfully, we make the following assumption for the network of participating agents.

\noindent
\rm\textbf{Assumption 4 (\underline{A4})}
\begin{itemize}
	\item The network of participating agents with topology $G$ is a strongly connected undirected graph.
\end{itemize}
\noindent
Note that the graph with undirected connections indicates that all the involved agents affect each other equally. Then, according to the consensus algorithm \cite{R2007Consensus}, we can obtain the following local interaction process of each agent
\begin{equation}
	g(\theta_{k}^{(i)}, e + 1) = g(\theta_{k}^{(i)}, e) + \epsilon \sum_{l \in \Omega_{i}}\left[g(\theta_{k}^{(l)}, e) - g(\theta_{k}^{(i)}, e)\right],
\end{equation}
where $\Omega_{i}$ represents the neighbors set of agent $i$. Considering the network of participating agents, $\Omega_{i}$ can be regarded as the set of the agents that are directly connected with agent $i$. $\epsilon$ denotes the local interaction step size, which plays a similar role as the learning rate $\eta$. Besides, $0 < \epsilon < 1/\Delta$, where $\Delta$ denotes the maximal degree of the graph and is defined by $\Delta := \max_i | \Omega_{i}| + 1$. Furthermore, the update rule of $\theta_{k}^{(i)}$ and $\bar{\theta}_{k}$ under the consensus-based method can be modified as
\begin{equation}
	\theta_{k}^{(i)} = \bar{\theta}_{t_0} - \eta \sum_{y = t_0}^{k - 1} g(\theta_{y}^{(i)}, e);
	\label{conagent}
\end{equation}
\begin{equation}
	\bar{\theta}_{k} = \bar{\theta}_{t_0} - \eta \frac{1}{m} \sum_{i = 1}^{m} \sum_{y = t_0}^{k - 1} g(\theta_{y}^{(i)}, e).
	\label{convirtual}
\end{equation}
The training algorithm of FMARL under the consensus-based method is illustrated in Algorithm 2. An intuitive schematic of the model's training process is presented in Fig. \ref{fig2}(c). It can be observed that agents need to exchange their local gradients with nearby collaborators before performing the local update, and these exchanged gradients are also used to update the model's average parameters. Note that the update rules in (\ref{conagent}) and (\ref{convirtual}) also conform to \underline{A2}, since the delay in obtaining local gradients can only affect the value of $g(\theta_{k}^{(i)}, 0)$, which equals to $\mathbf{0}$ for some agents at the beginning of local interaction. Thus, although local interactions occur synchronously between neighboring agents, it is not necessary for an agent to wait for all the neighbors to finish a step in Algorithm 2, so as to reduce the training time. Furthermore, we can obtain the following theorem.

\noindent
\rm\textbf{Theorem 2 (\underline{T2})} \textit{Suppose the number of local updates for agent $i$ is $\tau_i$, and the model's training process follows Algorithm 2. Under \underline{A1}, \underline{A2}, and \underline{A4}, if the total number of iterations $K$ is large enough and divisible by $\tau$, and the learning rate $\eta$ satisfies (\ref{t1}), then the expected gradient norm after $K$ iterations is bounded by}
\begin{align}
	\mathbb{E}\left[\frac{1}{K} \sum_{k=0}^{K-1} \left\|\nabla F(\bar{\theta}_k)\right\|^{2}\right] 
	\le  \underbrace{\frac{2[F(\bar{\theta}_0) - F_{\mathrm{inf}}]}{\eta K}  +  \frac{\eta L \sigma^{2}}{m}}  \notag \\
	\underbrace{+\ \eta^{2} \sigma^{2} L^{2}(\tau + 1)\left[1 - \epsilon \mu_{2}(\mathbf{La})\right]^{2E}}_{\psi_1^{(\text{C})}} \label{lb5},
\end{align}
where $E$ represents the total number of local interactions before each local update. $\mathbf{La}$ denotes the Laplace matrix of the graph $G$. $\mu_{2}(\mathbf{La})$ denotes the second smallest eigenvalue of $\mathbf{La}$, which is also called \textit{algebraic connectivity} \cite{R2007Consensus}. The corresponding proofs are presented in the Appendix F.

\noindent
\emph{Remarks:} Compared with the bound $\psi_1^{(\text{P})}$ obtained in (\ref{lb1}), we can find that the error convergence bound $\psi_1^{(\text{C})}$ under the consensus-based method is additionally affected by the topological properties (i.e., algebraic connectivity) of the graph comprised by agents and their connections. Thus, it is completely different from the consensus-based FL algorithm in \cite{Multi-Stage2020Seyyedali}, which utilizes an upper bound on the spectral radius of a stochastic matrix for interaction. Moreover, the conclusion in (\ref{lb5}) further considers the effect of local interaction step size $\epsilon$, which can be properly adjusted to guarantee the policy improvement of DRL \cite{Xu2022Trustable}. In addition, since $0 < \mu_{2}(\mathbf{La}) \le \Delta$ and the equality holds only when $G$ is a fully connected graph, we can find that $0 < 1 - \epsilon \mu_{2}(\mathbf{La}) < 1$. Therefore, we can infer that the implementation of local interactions can reduce the error convergence bound greatly. In addition, a larger step size $\epsilon$ or a more densely connected network of agents can also help reduce this bound. In practical applications, a small number of local interactions can make this bound decrease dramatically, such as $E = 2$.

\begin{algorithm}[htbp]
	\small
	\caption{\textbf{The training algorithm of FMARL under the consensus-based method.}}
	\begin{algorithmic}[1]
		\REQUIRE the model's initial parameters $\bar{\theta}_0$;
		\ENSURE the model's final average parameters $\bar{\theta}_k$;
		\STATE \textbf{Initialize} entire environment, learning rate $\eta$, loss function $\mathcal{L}$, reward function $\mathcal{R}$, maximal length of an epoch $T$, total number of epochs for training $U$, maximal size of a mini-batch $P$, number of local updates $\tau$ for agent $i=1$, number of agents that need to transmit the model's local gradients $m$, iteration index $k$, step size $\epsilon$, and total number of local interactions $E$;
		\FOR {epoch $u=1,2,3,...,U$}
		\FOR {transition $t=0,1,2,...,T-1$}
		\FOR {agent $i=1,2,3,...,m$}
		\STATE Calculate the input vector $s_t$ according to the received local state from the environment;
		\STATE Select an action $a_t$ according to $\theta_k^{(i)}(s_t,\ a_t)$;
		\STATE Perform the selected action and receive the next state $s_{t+1}$ from the environment;
		\STATE Calculate $r_t$ according to the reward function $\mathcal{R}$;
		\STATE Store this transition as $\phi_t^{(i)}$;
		\IF {$t+1$ mod $P$ == 0 \textbf{or} $t$ == $T-1$}
		\STATE Form the mini-batch $\xi_k^{(i)}$ by the stored transitions $\phi_t^{(i)}$;
		\STATE Calculate the mini-batch gradients by \\ $g(\theta_k^{(i)}) = \frac{1}{|\xi_k^{(i)}|} \sum_{\phi_{t}^{(i)} \in \xi_k^{(i)}} \nabla \mathbf{\mathcal{L}}(\theta_k^{(i)};\phi_{t}^{(i)})$;
		\STATE $g(\theta_{k}^{(i)}, 0) \gets g(\theta_k^{(i)})$;
		\STATE Wait for other agents;
		\FOR {interaction $e = 0,1,2,...,E-1$}
		\STATE $g(\theta_{k}^{(i)}, e + 1) = g(\theta_{k}^{(i)}, e) + \epsilon \sum_{l \in \Omega_{i}}\left[g(\theta_{k}^{(l)}, e) - g(\theta_{k}^{(i)}, e)\right]$;
		\ENDFOR 
		\STATE Perform the local update by \\ $\mathbf{\theta}_{k+1}^{(i)} = \mathbf{\theta}_{k}^{(i)} - \eta g(\theta_k^{(i)}, E)$;
		\STATE Clear the stored transitions $\phi_t^{(i)}$;
		\STATE Store the mini-batch gradients $g(\theta_k^{(i)}, E)$;
		\STATE $k \gets k+1$;
		\ELSE 
		\STATE $g(\theta_{k}^{(i)}, 0) \gets \mathbf{0}$;
		\ENDIF
		\IF {$k$ mod $\tau == 0$ $\textbf{or}$ reach the end of a period}
		\STATE Transmit all the accumulated gradients $g(\theta_y^{(i)}, E)$ to the virtual agent and receive the model's average parameters by \\
		$\bar{\theta}_{k} = \bar{\theta}_{t_0} - \eta \frac{1}{m} \sum_{i = 1}^{m} \sum_{y = t_0}^{k - 1} g(\theta_{y}^{(i)}, E)$;
		\STATE $\theta_k^{(i)} \gets \bar{\theta}_k$;
		\STATE Clear the stored transitions $\phi_t^{(i)}$;
		\STATE Clear the accumulated gradients $g(\theta_y^{(i)}, E)$;
		\ENDIF
		\ENDFOR
		\ENDFOR
		\ENDFOR 
		\STATE \textbf{Return} the model's average parameters $\bar{\theta}_k$;
	\end{algorithmic}
\end{algorithm}

On the other hand, regarding the resource cost under the consensus-based method, we assume that the extra communication overheads required for an agent to exchange the mini-batch gradients with one of its neighbors are $W_1$, and the computation overheads required to perform a local interaction are $W_2$. Then the system's resource cost can be reformulated from (\ref{psi0}) as
\begin{equation}
	\label{psi0_C}
	\psi_0^{(\text{C})} = \sum_{i=1}^{m} \left[\frac{C_1TU}{\tau P} + \frac{C_2 \tau_i TU}{\tau P} + |\Omega_{i}|(W_1 +W_2)\frac{ETU}{P}\right],
\end{equation}
where $|\Omega_{i}|$ denotes the size of $\Omega_{i}$. Here, the possible collision or interference issues in actual communication are omitted. Compared with $\psi_0$ defined in (\ref{psi0}), it can be observed from \eqref{psi0_C} that the additional local interactions under the consensus-based method increase the resource cost. However, since the model's error convergence bound is reduced at the same time, the system's utility value may be improved in certain cases. For example, suppose that the overheads required for participating agents to implement the D2D communication are much less than those to transmit the same messages to the remote virtual agent, the effectiveness of the consensus-based method would be clearly demonstrated. In addition, participating agents can intentionally set up as a sparse graph (i.e., smaller $|\Omega_{i}|$) to decrease the number of local interactions between agents, so as to further reduce the resource cost.

\begin{figure}
	\centering
	\includegraphics[width=0.4\textwidth]{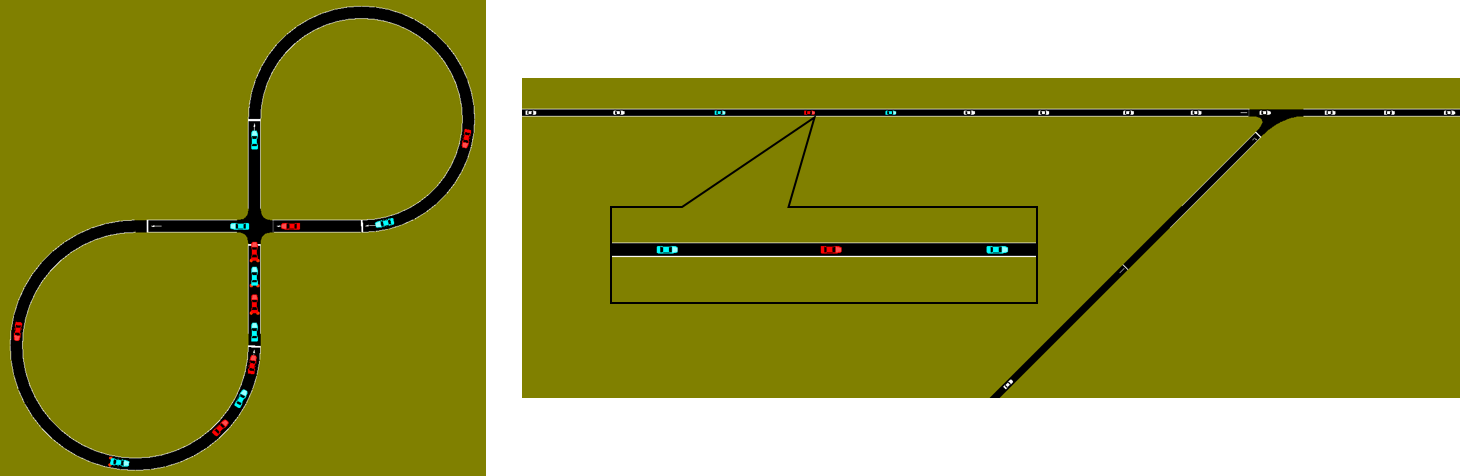}
	\caption{Figure Eight (left) and Merge (right).}
	\label{fig3}
\end{figure}

\section{Simulation Results}
\label{sec:result}
In this section, we provide the simulation results about the two proposed optimization methods. The simulation scenarios are taken from \cite{VinitskyBenchmarks}, which is released as a new benchmark in traffic control through creating DRL controllers for mixed-autonomy traffic. As illustrated in Fig. \ref{fig3}, two simulation scenarios in this benchmark are selected to verify the effectiveness and efficiency of the developed methods. For the first scenario ``Figure Eight", there are totally 14 vehicles running circularly along a one-way lane that resembles the shape of ``8". An intersection is located at the lane, and each vehicle must control its acceleration to pass through this intersection, so as to increase the average speed of the whole vehicle team. Note that slamming on the brakes will be forced on the vehicles that are about to crash, and the current epoch will be terminated once the collision occurs. We further modify the ``Figure Eight" scenario to assign the related local state to each vehicle, including its position and speed as well as those of the vehicle ahead and behind. Depending on its local state, each vehicle needs to optimize its acceleration, which is a normalized continuous variable  between $-1$ and $1$. All the involved vehicles are the same, and unless additional controllers are assigned, these vehicles are controlled by the underlying simulation of urban mobility (SUMO) in the same mode, which is a widely used and open-source traffic simulation package \cite{VinitskyBenchmarks}. In the ``Figure Eight" scenario, half of the vehicles are empowered by the DRL-based controllers.

The simulation settings of the second scenario ``Merge" are almost the same as those of the first scenario, except that the maximal speed and acceleration of each vehicle are increased. The ``Merge" scenario simulates the intersection of a highway and a side lane, and each vehicle needs to control its acceleration to increase the average speed. Moreover, there are totally $50$ vehicles in the second scenario, and $5$ vehicles are randomly selected within each epoch to instantiate the DRL-based controllers. Note that we take the normalized average speed (NAS) of all vehicles at each iteration as the individual reward in each scenario, which is assigned to each training vehicle after its action being performed. Unless otherwise specified, the DRL-based controllers for each scenario are optimized through the PPO algorithm \cite{John2017Proximal}. Main parameters in the simulations are listed in Table \ref{tb3}.

\begin{table}[tbp]
	\footnotesize
	\centering
	\caption{\label{tb3}Main parameters in the simulations.}
	\begin{tabular}{c|c}
		\toprule[1.1pt]
		Parameters                                     & Value     \\
		\hline	
		Number of agents, m                            & 7 (or 5)  \\     
		Learning rate, $\eta$                          & $10^{-4}$ \\
		Length of epoch, T                             & 1500      \\   
		Number of epochs, U                            & 500       \\    
		Mini-batch size, P                             & 250       \\   
		Number of iterations, K                        & 3000      \\	  
		Maximal local updates, $\tau$                  & 15        \\     
		Step size in local interaction, $\epsilon$     & 0.1       \\ 
		Discount factor in PPO \cite{John2017Proximal} & 0.9       \\
		Clipping parameter in PPO                      & 0.2       \\
		Epoch parameter in PPO                         & 4         \\
		\bottomrule[1.1pt]
	\end{tabular}
\end{table}

\begin{figure}
	\centering
	\includegraphics[width=0.4\textwidth]{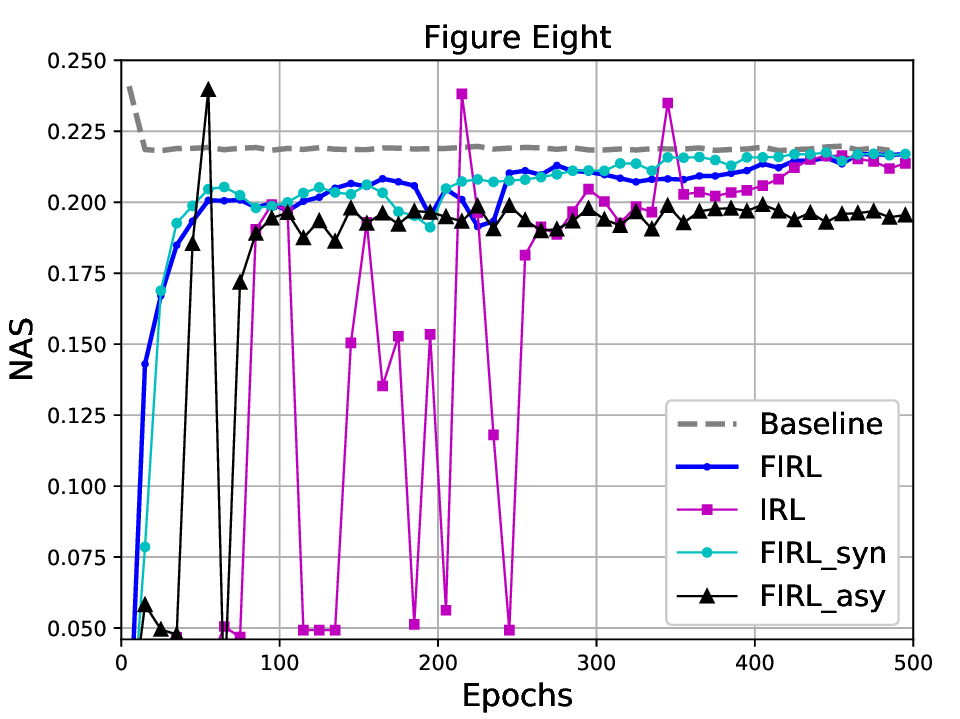}
	\caption{Effectiveness of FL in IRL.}
	\label{figa}
\end{figure}

In Fig. \ref{figa}, we first present the simulation performance of naive \texttt{IRL} and \texttt{FIRL} (\texttt{IRL} with FL). Here, NAS denotes the normalized average speed of all vehicles during an entire testing epoch. The test is performed every 10 epochs and takes the average of five repetitions. We further take the performance of all vehicles controlled by the underlying SUMO as the optimal baseline. Fig. \ref{figa} shows that FL can clearly improve the performance of IRL in terms of training efficiency and stability, while verify the necessity of this combination. Meanwhile, we test the effect of delay on \texttt{FIRL} in the method \texttt{FIRL\_syn} and \texttt{FIRL\_asy}. Specifically, both methods introduce the delay of 2 epochs (i.e., 300 seconds) to the up-link and down-link between each agent and the virtual agent. However, agents in \texttt{FIRL\_syn} can update their parameters every iteration, while agents in \texttt{FIRL\_asy} can only update their parameters every 2 epochs, which implies a worsening communication. As indicated in Fig. \ref{figa}, a smaller updating frequency may reduce the effectiveness of FL in IRL.

\begin{figure}[htbp]
	\begin{center}
		\begin{minipage}{0.75\linewidth}
			\centerline{\includegraphics[width=1\linewidth]{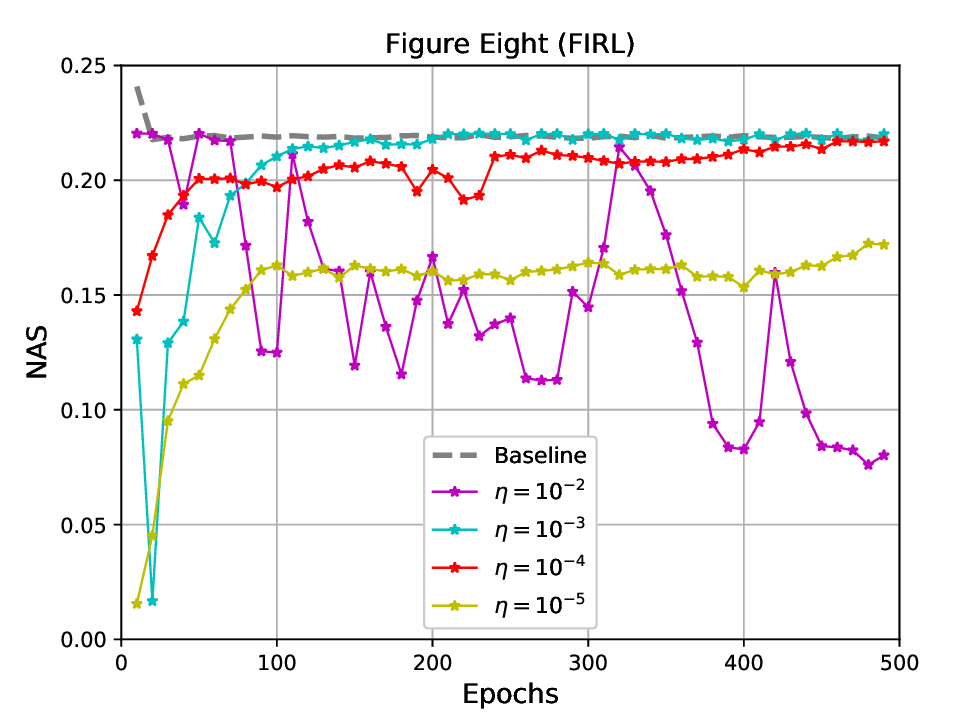}}
			\centerline{\small{(a)}}
		\end{minipage}
	\end{center}
	\vfill
	\begin{center}
		\begin{minipage}{0.75\linewidth}
			\centerline{\includegraphics[width=1\linewidth]{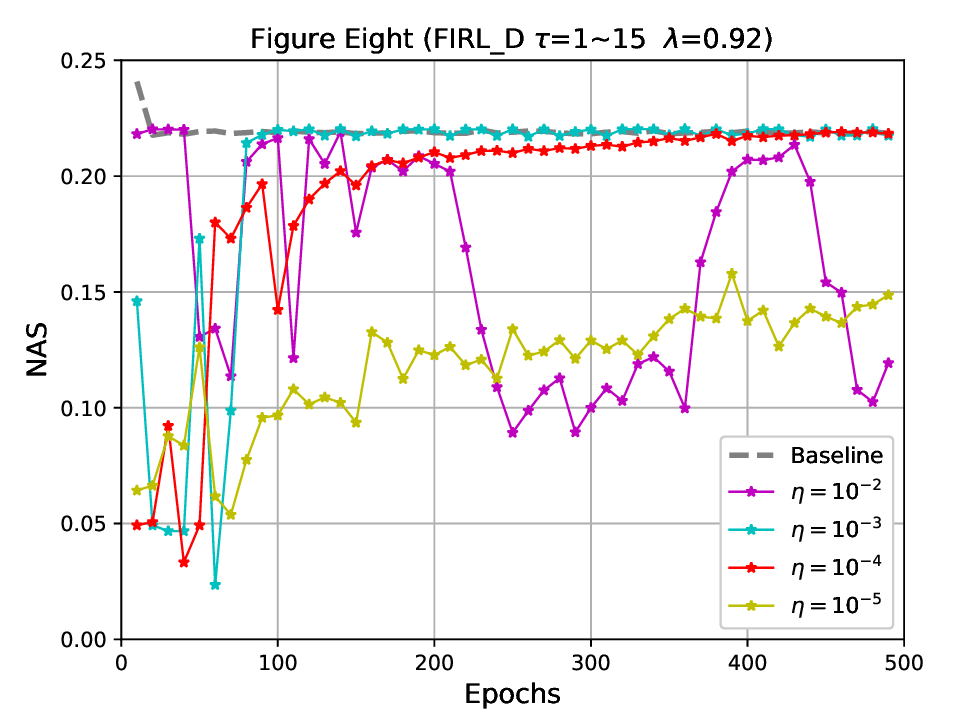}}
			\centerline{\small{(b)}}
		\end{minipage}
	\end{center}
	\caption{Performance of (a) \texttt{FIRL} and (b) \texttt{FIRL\_D} under different $\eta$.}
	\label{fig9}
\end{figure}

Besides, we provide the performance of \texttt{FIRL} and \texttt{FIRL\_D} with $\tau=1\sim 15$, $\lambda=0.92$ under different $\eta$ in Fig. \ref{fig9}. Note that the total number of policy iterations $K = UT/P$ and $K$ is proportional to the number of epochs $U$. We can observe from Figs. \ref{fig9}(a) and (b) that with a proper $\eta$ (e.g., $\eta = 10^{-3}$), the performance gradually converges with the increase of $K$. However, an over-trivial $\eta$ may slow the convergence rate, as shown by the curves with $\eta=10^{-5}$. On the other hand, since the model's gradients are calculated by the loss function in DRL, an overlarge step size of $\eta$ would suffer from the catastrophic forgetting problem and make the model hard to converge \cite{Kakade2002Approximately}, as shown by the curves with $\eta=10^{-2}$. Therefore, the selection of $\eta$ should balance the trade-off between convergence rate and convergence performance.

\begin{figure*}
	\begin{minipage}{0.32\linewidth}
		\centerline{\includegraphics[width=1\linewidth]{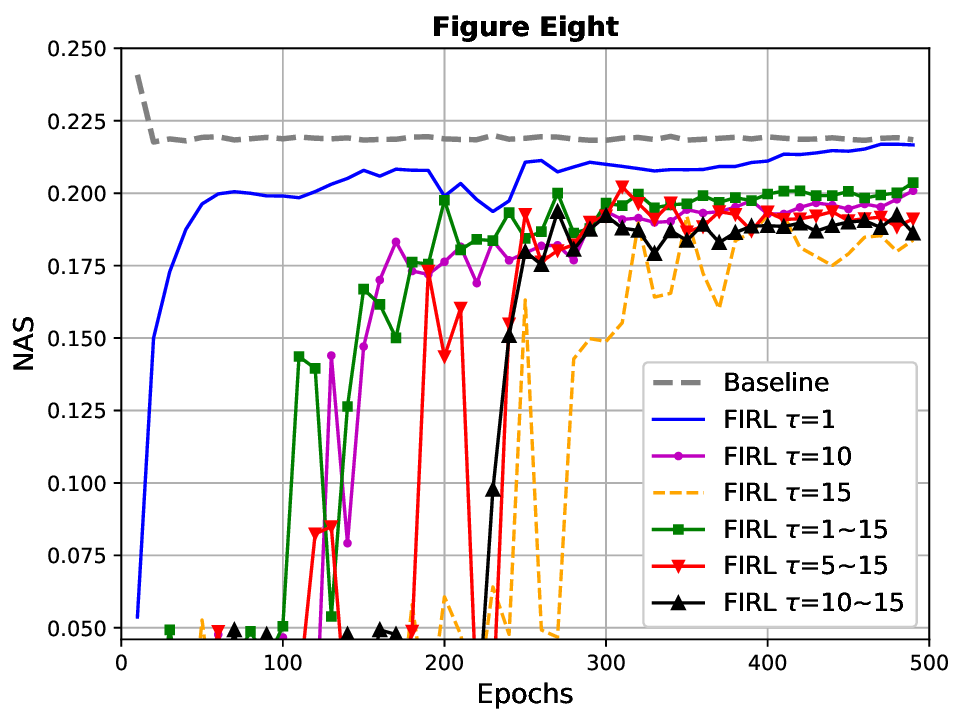}}
		\centerline{\small{(a)}}
	\end{minipage}
	\hfill
	\begin{minipage}{.32\linewidth}
		\centerline{\includegraphics[width=1\linewidth]{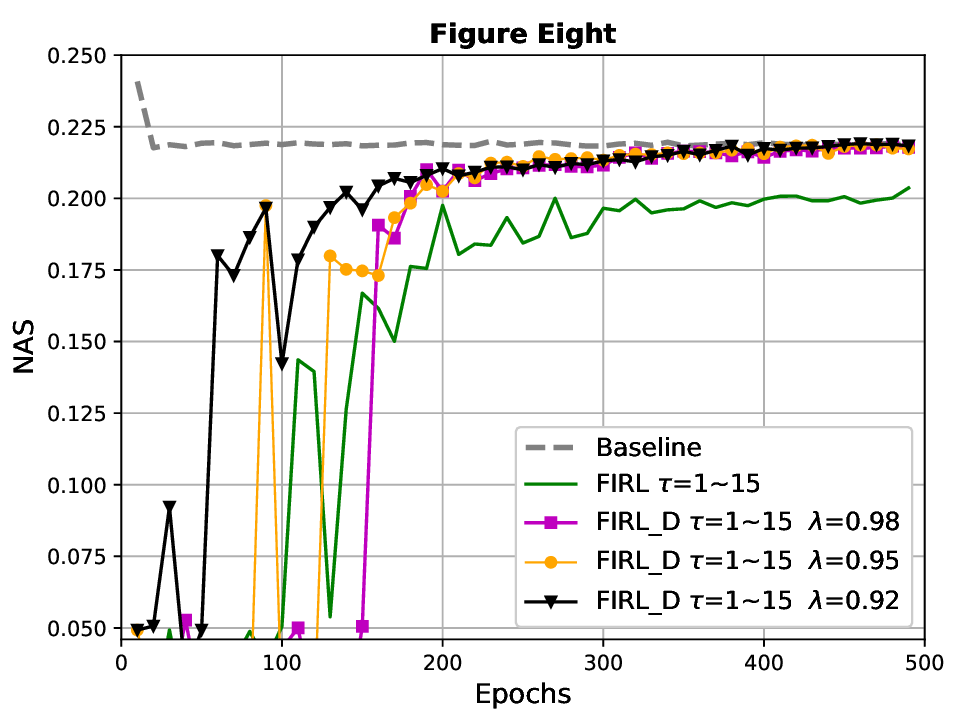}}
		\centerline{\small{(b)}}
	\end{minipage}
	\hfill
	\begin{minipage}{.32\linewidth}
		\centerline{\includegraphics[width=1\linewidth]{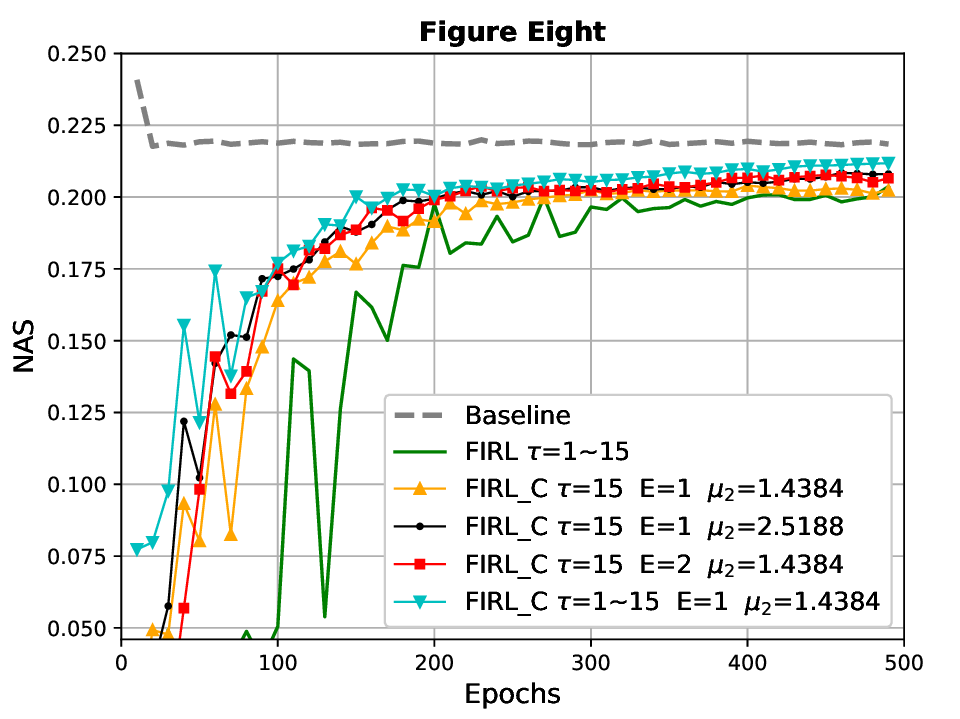}}
		\centerline{\small{(c)}}
	\end{minipage}
	\vfill
	\begin{minipage}{0.32\linewidth}
		\centerline{\includegraphics[width=1\linewidth]{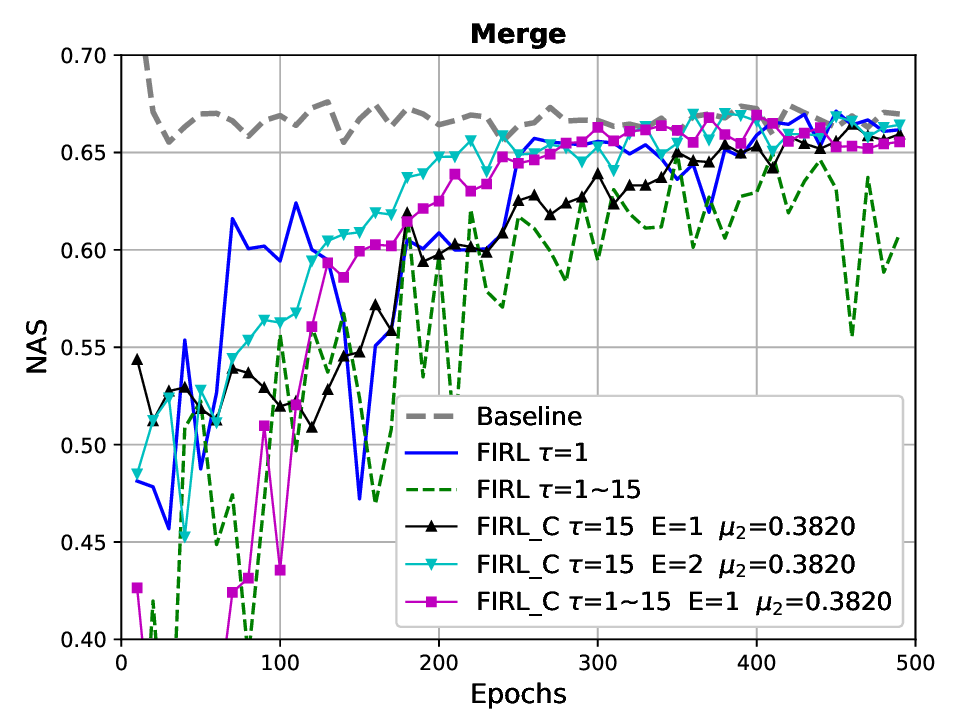}}
		\centerline{\small{(d)}}
	\end{minipage}
	\hfill
	\begin{minipage}{.32\linewidth}
		\centerline{\includegraphics[width=1\linewidth]{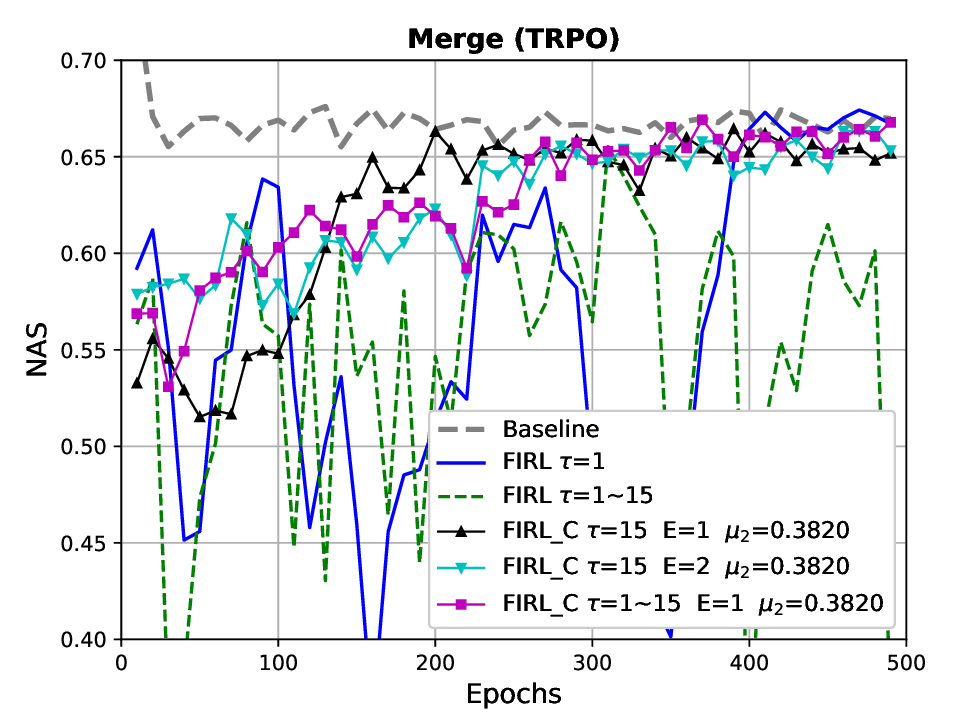}}
		\centerline{\small{(e)}}
	\end{minipage}
	\hfill
	\begin{minipage}{.32\linewidth}
		\centerline{\includegraphics[width=1\linewidth]{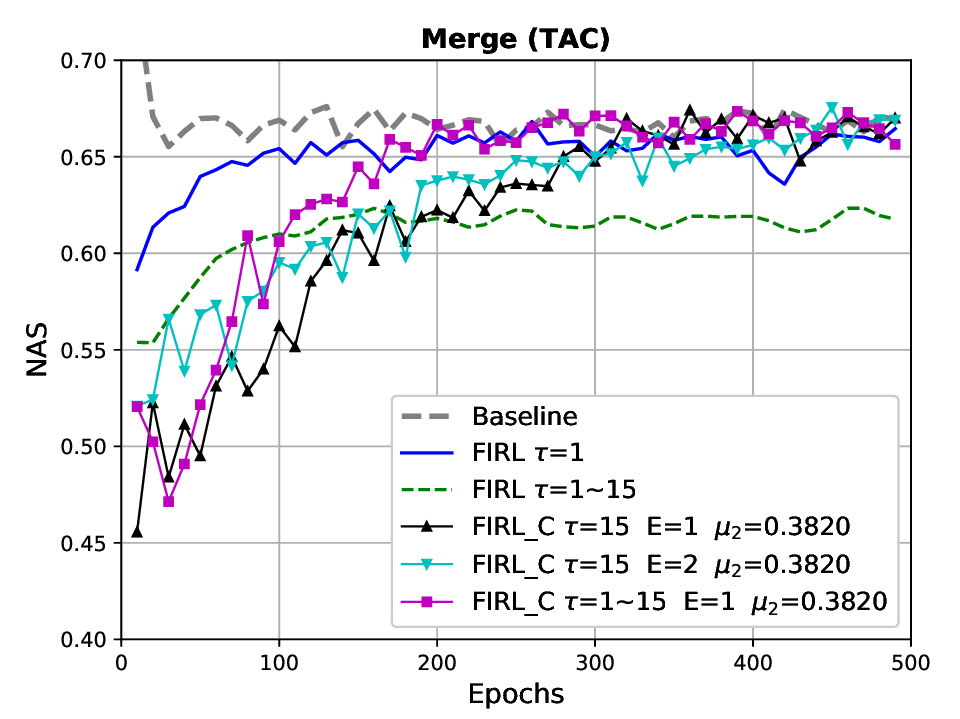}}
		\centerline{\small{(f)}}
	\end{minipage}
	\caption{Convergence performance of (a) the VPA method, (b) the decay-based method, (c) the consensus-based method, (d) the PPO algorithm, (e) the TRPO algorithm, and (f) the TAC algorithm.}
	\label{fig4}
\end{figure*}

In Fig. \ref{fig4}(a), we present the performance of \texttt{FIRL} with different local updates in a period. Here, the notation ``$\tau=1\sim15$" denotes that the numbers of local updates from agents are uniformly distributed between $1$ and $15$, thus $\nu=8$. We set the maximal value of $\tau$ to $15$, so as to highlight the performance improvement of the proposed methods on the classical periodic averaging method. It can be observed from Fig. \ref{fig4}(a) that the performance will decline as the number of local updates $\tau$ or its mean value $\nu$ increases, which is consistent with the conclusion in (\ref{lb2}). In Fig. \ref{fig4}(b), we present the performance of \texttt{FIRL} under the decay-based method (i.e., \texttt{FIRL\_D}), which realizes the practical implementation in \underline{C1} when the numbers of local updates from agents are uniformly distributed between $1$ and $15$. Fig. \ref{fig4}(b) demonstrates that \texttt{FIRL\_D} can improve the model's convergence performance, and a smaller decay constant $\lambda$ generates a faster convergence rate, which are consistent with our previous discussions in \underline{T1} and \underline{C1}. In Fig. \ref{fig4}(c), we present the performance of \texttt{FIRL} under the consensus-based method (i.e., \texttt{FIRL\_C}). On the premise of strong connectivity, the topology of agents' network with the algebraic connectivity $\mu_{2}=1.4384$ is constructed by $3 \sim 4$ random connections from each agent to nearby collaborators, while these connections are increased to $4 \sim 6$ when $\mu_{2}=2.5188$. Note that there is only one connection between any pair of agents. It can be observed from Fig. \ref{fig4}(c) that the model's convergence performance is improved when the local interactions are considered, even when various numbers of local updates are introduced, which are consistent with the conclusions in \underline{T2}. In addition, the topology with either a larger $\mu_{2}$ (i.e., more connections) or a greater $E$ (i.e., more local interactions) generally obtains better performance. In Figs. \ref{fig4}(d) - (f), we further present the performance of \texttt{FIRL\_C} in the ``Merge" scenario under different policy gradient methods. Specifically, since vehicles in the ``Merge" scenario are running along a highway, we construct the topology of agents' network by connecting any two adjacent DRL-based vehicles, thus $\mu_{2}=0.3820$. Besides, we use the trust region policy optimization (TRPO) method \cite{John2015Trust} and the Tsallis actor-critic (TAC) algorithm \cite{Tsallis2019Kyungjae} as the corresponding policy gradient method in Fig. \ref{fig4}(e) and Fig. \ref{fig4}(f), respectively. We can observe from Figs. \ref{fig4}(d) - (f) that \texttt{FIRL\_C} can stabilize the model's training process, meanwhile performs well across different policy gradient methods.  

\begin{figure}[htbp]
	\begin{center}
		\begin{minipage}{0.75\linewidth}
			\centerline{\includegraphics[width=1\linewidth]{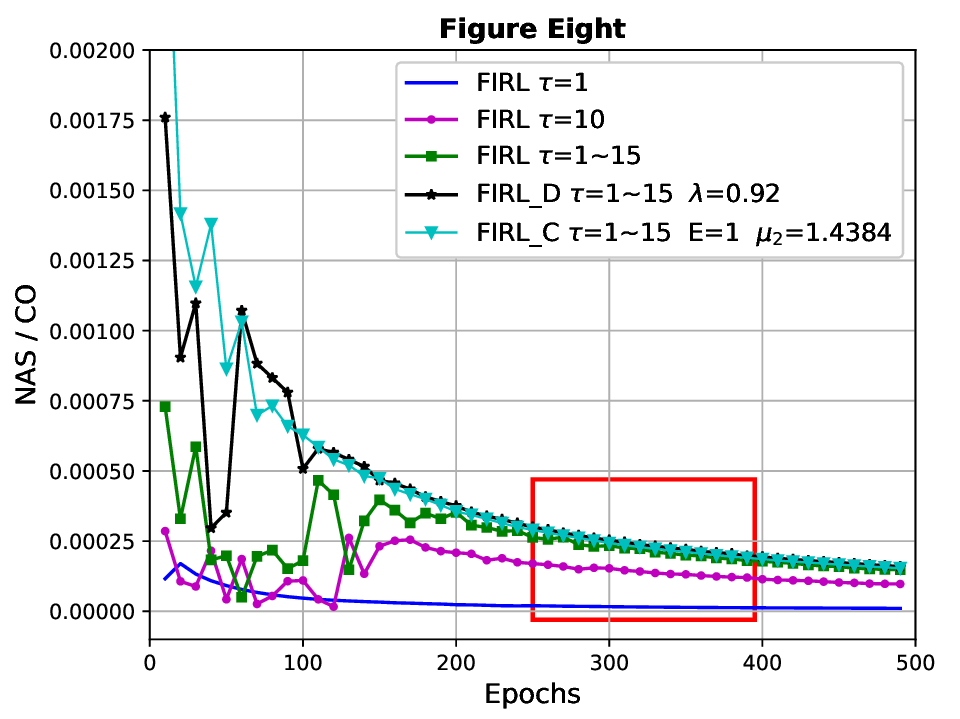}}
			\centerline{\small{(a)}}
		\end{minipage}
	\end{center}
	\vfill
	\begin{center}
		\begin{minipage}{0.75\linewidth}
			\centerline{\includegraphics[width=1\linewidth]{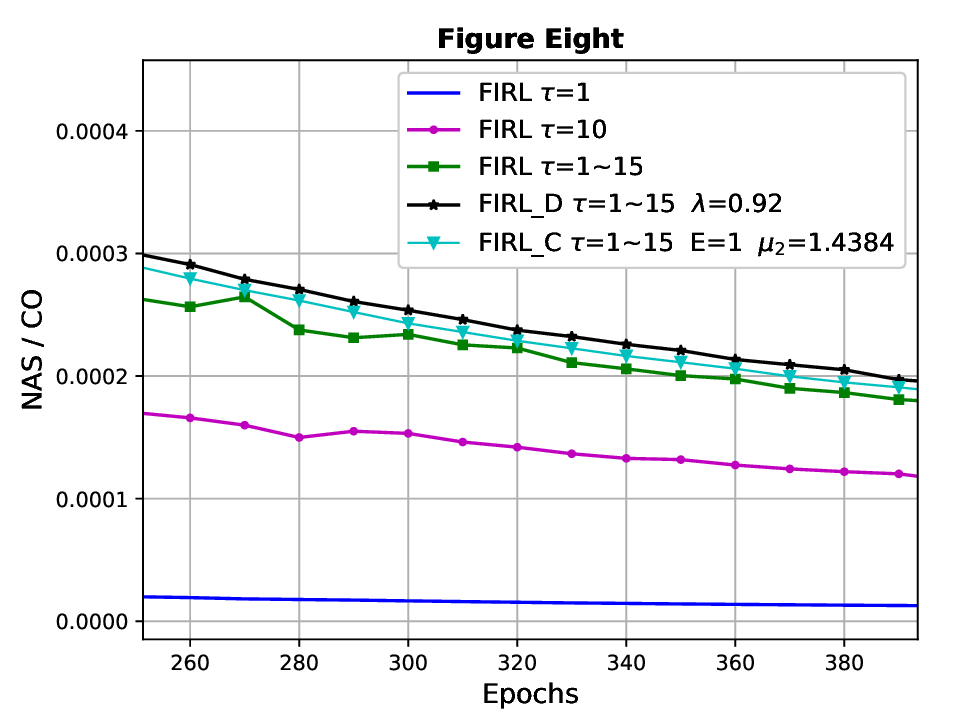}}
			\centerline{\small{(b)}}
		\end{minipage}
	\end{center}
	\caption{(a) Ratio of NAS to communication overhead (CO) with respect to epochs in the ``Figure Eight" scenario, and (b) magnified image in the red box of (a).}
	\label{fig5}
\end{figure}

In Fig. \ref{fig5}, we select some representative methods in Figs. \ref{figa} and \ref{fig4} to indicate the ratio of NAS to communication overhead (CO) with respect to epochs. Here, the ratio of NAS to CO denotes the performance improvement per unit of communication overhead and indicates the communication efficiency, where the parameter $C_1$ is set to $1$ and $W_1$ is set to $1.0 \times 10^{-3}C_1$. A more detailed description about the communication overheads of these methods is provided in Table \ref{tb2}. It can be observed from Fig. \ref{fig5}(a) that compared with \texttt{FIRL} with $\tau=1$, $\tau=10$, and $\tau=1 \sim 15$, \texttt{FIRL\_D} and \texttt{FIRL\_C} have higher starting points, which means their performance (i.e., NAS) increases faster when the communication overheads are small. Since the value of CO under this parameter setting is much greater than that of NAS, the methods with the same maximal range of $\tau$ converge to a similar level, such as \texttt{FIRL} with $\tau=1 \sim 15$, \texttt{FIRL\_D}, and \texttt{FIRL\_C}. However, Fig. \ref{fig5}(b), a magnified image of the red box in Fig. \ref{fig5}(a), clearly shows that the proposed method \texttt{FIRL\_D} and \texttt{FIRL\_C} maintain better performance.

\begin{table*}[htbp]
	\scriptsize
	\centering
	\caption{Numerical simulation results in the ``Figure Eight" scenario.}
	\label{tb2}
	\begin{tabular}{c|c|c|c|c|c|c|c}
		\toprule[1.1pt]
		\multirow{2}{*}{Methods} & \multirow{2}{*}{Local updates} & \multirow{2}{*}{Decay-based} & \multirow{2}{*}{Consensus-based} & Communication             & Computation                & Expected      & Normalized    \\
		~                        & ~                              & ~                            & ~                                & overheads                 & overheads                  & gradient norm & utility value \\
		\hline
		\texttt{FIRL}            & $\tau = 1$                     & None                         & None                             & 21000 $C_1$               & 21000 $C_2$                & 1.5590        & 0.0           \\
		\texttt{FIRL}            & $\tau = 10$                    & None                         & None                             & 2100 $C_1$                & 21000 $C_2$                & 6.3421        & 0.5734        \\
		\texttt{FIRL}            & $\tau = 15$                    & None                         & None                             & 1400 $C_1$                & 21000 $C_2$                & 9.6069        & 0.7809        \\
		\texttt{FIRL}            & $\tau = 10\thicksim 15$        & None                         & None                             & 1400 $C_1$                & 19000 $C_2$                & 10.1892       & 0.7601        \\
		\texttt{FIRL}            & $\tau = 5\thicksim15$          & None                         & None                             & 1400 $C_1$                & 16200 $C_2$                & 8.7182        & 0.8130        \\
		\texttt{FIRL}            & $\tau = 1\thicksim15$          & None                         & None                             & 1400 $C_1$                & 12600 $C_2$                & 7.6476        & 0.8516        \\
		\texttt{FIRL\_D}         & $\tau = 1\thicksim15$          & $\lambda = 0.98$             & None                             & 1400 $C_1$                & 12600 $C_2$                & 7.2782        & 0.8649        \\
		\texttt{FIRL\_D}         & $\tau = 1\thicksim15$          & $\lambda = 0.95$             & None                             & 1400 $C_1$                & 12600 $C_2$                & 7.2537        & 0.8657        \\
		\texttt{FIRL\_D}         & $\tau = 1\thicksim15$          & $\lambda = 0.92$             & None                             & 1400 $C_1$                & 12600 $C_2$                & 3.5090        & 1.0           \\
		\texttt{FIRL\_C}         & $\tau = 15$                    & None                         & $E=1$, $\mu_{2}=1.4384$          & 1400 $C_1$ + 78000 $W_1$  & 21000 $C_2$ + 78000 $W_2$  & 3.6188        & 0.9336        \\
		\texttt{FIRL\_C}         & $\tau = 15$                    & None                         & $E=1$, $\mu_{2}=2.5188$          & 1400 $C_1$ + 96000 $W_1$  & 21000 $C_2$ + 96000 $W_2$  & 2.1648        & 0.9688        \\
		\texttt{FIRL\_C}         & $\tau = 15$                    & None                         & $E=2$, $\mu_{2}=1.4384$          & 1400 $C_1$ + 156000 $W_1$ & 21000 $C_2$ + 156000 $W_2$ & 2.8746        & 0.9023        \\
		\texttt{FIRL\_C}         & $\tau = 1\thicksim15$          & None                         & $E=1$, $\mu_{2}=1.4384$          & 1400 $C_1$ + 78000 $W_1$  & 12600 $C_2$ + 78000 $W_2$  & 3.6072        & 0.9346        \\
		\bottomrule[1.1pt]
	\end{tabular}
\end{table*}

In Table \ref{tb2}, we summarize the numerical simulation results in the ``Figure Eight" scenario, so as to verify the effectiveness of the developed methods in improving the system's utility value. In particular, the expected gradient norm is calculated upon a predetermined sample set, which is comprised by samples that are uniformly collected during the model's training process when $\tau = 1$. Moreover, the expected gradient norm is calculated whenever the model's average parameters (i.e., $\bar{\theta}_{k}$) are updated, and its final value is averaged across the entire training process. In Table \ref{tb2}, with the same communication and computation overheads as \texttt{FIRL} with $\tau = 1 \sim 15$, \texttt{FIRL\_D} maintains a smaller expected gradient norm, thus having a greater system's utility value. Besides, compared with \texttt{FIRL} with $\tau = 15$ or $\tau = 1 \sim 15$, \texttt{FIRL\_C} requires the same amount of resource cost in FL, and although there are more resources required in local interactions, its expected gradient norm is smaller. Thus, the system's utility value can be also improved when $W_1$ and $W_2$ are small in \texttt{FIRL\_C}. As an example, we give the system's normalized utility value in Table \ref{tb2} under the conditions $C_1=1$, $W_1=1.0 \times 10^{-3}C_1$, $C_2 = 1.0 \times 10^{-4}C_1$, and $W_2=C_2$. We can observe that compared with \texttt{FIRL}, \texttt{FIRL\_D} and \texttt{FIRL\_C} maintain higher utility values. Specifically, the utility value of \texttt{FIRL\_C} will increase as the computation overheads decrease.

\section{Conclusions}
\label{sec:conclusion}
This paper has taken advantage of the paradigm of FL to improve the policy performance of IRL agents. Meanwhile, considering the excessive communication overheads generated between agents and a central server in FL as well as the heterogeneity of independent learning environments faced by IRL agents, this paper has built the framework of FMARL on the basis of the VPA method. Moreover, to reach a good balance between the system's resource cost and the model's convergence performance, a novel utility function has been proposed to quantify the convergence bound of the model's error reduction per unit of resource cost. Furthermore, to improve the system's utility value, we have put forward two new optimization methods on top of the VPA method. By analyzing the theoretical convergence bounds and performing extensive simulations, both effectiveness and efficiency of the developed methods have been verified. 

For future works, we plan to implement the proposed optimization methods in the real-world applications. In practice, when there are a large number of participating agents, multiple virtual agents may emerge simultaneously and their organization tends to be hierarchical, which means a more complex scenario and requires more careful considerations on the optimization methods. Moreover, we also plan to optimize the aggregation weights for agents with different sample sizes in FL, which promises an effective method to improve the system's performance. Besides, it is also interesting to develop more smart means to determine the hyperparameter (e.g., $\eta$). Finally, whether the common shared learning model can be applied to heterogeneous tasks faced by different agents remains an open question that is worth exploring. 

\section*{Acknowledgement}
This work was supported in part by the National Natural Science Foundation of China under Grants 62071425, in part by the Zhejiang Key Research and Development Plan under Grant 2022C01093, in part by Huawei Cooperation Project, and in part by the Zhejiang Provincial Natural Science Foundation of China under Grant LR23F010005.

\begin{appendices}

	\section{Proof Preliminaries}
	In this section, we introduce some of the definitions and notations used across the Supplemental Material. In addition, we also introduce several important lemmas, on which the proofs of the proposed theorems are built. In particular, we define the average mini-batch gradients
	\begin{equation}
		\label{eq:minibatch_gradients}
		\mathcal{G}_k = \frac{1}{m} \sum_{i=1}^{m} g(\theta_{k}^{(i)}), 
	\end{equation}
	and the average full-batch gradients
	\begin{equation}
		\label{eq:fullbatch_gradients}
		\mathcal{H}_k = \frac{1}{m} \sum_{i=1}^{m} \nabla F(\theta_{k}^{(i)}).
	\end{equation}
	According to Condition 3 in \underline{A1}, we can obtain
	\begin{equation}
		\mathbb{E}[\mathcal{G}_k] = \mathcal{H}_k.
		\label{gk=hk}
	\end{equation}
	In addition, we define the Frobenius norm for $\mathbf{A} \in M_n$ by
	\begin{equation}
		\| \mathbf{A} \|_{\mathrm{F}}^2 = |\mathrm{Tr}(\mathbf{A}\mathbf{A}^{\top})| = \sum_{i=1}^{n} \sum_{j=1}^{n} | a_{i,j} |^2.
	\end{equation}
	And the operator norm for $\mathbf{A} \in M_n$ is defined by
	\begin{equation}
		\label{eq:operator_norm}
		\| \mathbf{A} \|_{\mathrm{op}} = \max_{\|x\|=1}\|\mathbf{A}x\| = \sqrt{\lambda_{\mathrm{max}}(\mathbf{A}^{\top}\mathbf{A})},
	\end{equation}
	where $\lambda_{\mathrm{max}}$ denotes the maximal eigenvalue. Then we can infer that for real matrix $\mathbf{A} \in \mathbb{R}^{d \times m}$ and $\mathbf{B} \in \mathbb{R}^{m \times m}$, if $\mathbf{B}$ is symmetric, then the following inequality holds
	\begin{equation}
		\label{eq:operator_frobenius_norm_inequality}
		\| \mathbf{A}\mathbf{B} \|_{\mathrm{F}} \le \| \mathbf{B} \|_{\mathrm{op}} \| \mathbf{A} \|_{\mathrm{F}}.
	\end{equation}
	Besides, we use the notation $\mathbf{1}$ to represent the vector $[1,1,1,...,1]_{1\times m}^{\top}$, and $\mathbf{J} := \mathbf{1}\mathbf{1}^{\top}/(\mathbf{1}^{\top} \mathbf{1}$).
	
	\vspace{1.0em}
	\noindent
	\rm\textbf{Lemma 1.} \textit{Under Condition 3 and 4 in \underline{A1}, the variance of the average mini-batch gradients is bounded by}
	\begin{equation}
		\mathbb{E} \left\| \mathcal{G}_k - \mathcal{H}_k \right\|^{2} \le \frac{\beta}{m^2} \sum_{i=1}^{m} \left\| \nabla F(\theta_{k}^{(i)}) \right\|^2 + \frac{\sigma^2}{m}.
	\end{equation}
	\begin{proof}
		According to \eqref{eq:minibatch_gradients} and \eqref{eq:fullbatch_gradients}, we have
		\begin{align}
			          & \mathbb{E} \| \mathcal{G}_k - \mathcal{H}_k \|^{2} = \mathbb{E} \left\| \frac{1}{m} \sum_{i=1}^{m} \left[g(\theta_{k}^{(i)}) - \nabla F(\theta_{k}^{(i)})\right] \right\|^{2} \notag                                                          \\
			          & = \frac{1}{m^{2}} \sum_{i=1}^{m} \mathbb{E} \left\| g(\theta_{k}^{(i)}) - \nabla F(\theta_{k}^{(i)}) \right\|^{2} \notag                                                                                                                      \\
			\nonumber & + \frac{1}{m^{2}} \mathbb{E} \left[\sum_{s \ne l}^{m} \left\langle g(\theta_{k}^{(s)}) - \nabla F(\theta_{k}^{(s)}), g(\theta_{k}^{(l)}) - \nabla F(\theta_{k}^{(l)}) \right\rangle\right]                                                    \\
			          & = \frac{1}{m^{2}} \sum_{i=1}^{m} \mathbb{E} \left\| g(\theta_{k}^{(i)}) - \nabla F(\theta_{k}^{(i)}) \right\|^{2} \notag                                                                                                                      \\
			\nonumber & + \frac{1}{m^{2}} \sum_{s \ne l}^{m} \left\langle \mathbb{E}_{\xi_k^{(s)}|\theta_{k}^{(s)}}\left[g(\theta_{k}^{(s)}) - \nabla F(\theta_{k}^{(s)})\right], \mathbb{E}_{\xi_k^{(l)}|\theta_{k}^{(l)}}\left[ g(\theta_{k}^{(l)}) \right. \right. \\
			\nonumber & \left. \left. - \nabla F(\theta_{k}^{(l)})\right] \right\rangle                                                                                                                                                                               \\
			          & \stackrel{(a)}{=} \frac{1}{m^{2}} \sum_{i=1}^{m} \mathbb{E} \left\| g(\theta_{k}^{(i)}) - \nabla F(\theta_{k}^{(i)}) \right\|^{2} \notag                                                                                                      \\
			          & \stackrel{(b)}{\le} \frac{1}{m^{2}} \sum_{i=1}^{m} \left[\beta \left\|\nabla F(\theta_{k}^{(i)})\right\|^2 + \sigma^2\right] \notag                                                                                                           \\
			          & =\frac{\beta}{m^2} \sum_{i=1}^{m} \left\| \nabla F(\theta_{k}^{(i)}) \right\|^2 + \frac{\sigma^2}{m}, \notag
		\end{align}
		where $\{\xi_k^{(s)}\}$ and $\{\xi_k^{(l)}\}$ are independent random variables. The equality $(a)$ is due to that according to Condition 3 in \underline{A1}, $\mathbb{E}_{\xi_k^{(s)}|\theta_{k}^{(s)}}\left[g(\theta_{k}^{(s)}) - \nabla F(\theta_{k}^{(s)})\right]$ and $\mathbb{E}_{\xi_k^{(l)}|\theta_{k}^{(l)}}\left[ g(\theta_{k}^{(l)}) - \nabla F(\theta_{k}^{(l)})\right]$ turn to be $0$. The inequality $(b)$ comes from Condition 4 in \underline{A1}.	
	\end{proof}
	
	\vspace{1.0em}
	\noindent
	\rm\textbf{Lemma 2.} \textit{Under Condition 3 in \underline{A1}, the expected inner product between the average mini-batch gradients and the full-batch gradients satisfies}
	\begin{align}
		  & \mathbb{E}\left[\left\langle \nabla F(\bar{\theta}_k), \mathcal{G}_k \right\rangle\right] \nonumber                             \\
		= & \frac{1}{2} \left\|\nabla F(\bar{\theta}_k)\right\|^2 + \frac{1}{2m} \sum_{i=1}^{m} \left\|\nabla F(\theta_{k}^{(i)})\right\|^2 \\
		  & - \frac{1}{2m} \sum_{i=1}^{m} \left\| \nabla F(\bar{\theta}_k) - \nabla F(\theta_{k}^{(i)}) \right\|^2.\nonumber 
	\end{align}
	\begin{proof}
		According to the definition of $\mathcal{G}_k$, we have
		\begin{align}
			          & \mathbb{E} \left[\left\langle \nabla F(\bar{\theta}_k), \mathcal{G}_k \right\rangle \right] = \mathbb{E} \left[\left\langle \nabla F(\bar{\theta}_k), \frac{1}{m} \sum_{i=1}^{m} g(\theta_{k}^{(i)}) \right\rangle \right] \notag \\
			          & \stackrel{(a)}{=} \left\langle \nabla F(\bar{\theta}_k), \frac{1}{m} \sum_{i=1}^{m} \nabla F(\theta_{k}^{(i)}) \right\rangle \notag                                                                                               \\
			          & = \frac{1}{m} \sum_{i=1}^{m} \left\langle \nabla F(\bar{\theta}_k), \nabla F(\theta_{k}^{(i)}) \right\rangle \notag                                                                                                               \\
			          & \stackrel{(b)}{=} \frac{1}{2m} \sum_{i=1}^{m} \left[\left\|\nabla F(\bar{\theta}_k)\right\|^2 + \left\|\nabla F(\theta_{k}^{(i)})\right\|^2 \right. \notag                                                                        \\
			\nonumber & \left. - \left\| \nabla F(\bar{\theta}_k) - \nabla F(\theta_{k}^{(i)}) \right\|^2 \right] \notag                                                                                                                                  \\
			          & = \frac{1}{2} \left\|\nabla F(\bar{\theta}_k)\right\|^2 + \frac{1}{2m} \sum_{i=1}^{m} \left\|\nabla F(\theta_{k}^{(i)})\right\|^2 \notag                                                                                          \\
			\nonumber & - \frac{1}{2m} \sum_{i=1}^{m} \left\| \nabla F(\bar{\theta}_k) - \nabla F(\theta_{k}^{(i)}) \right\|^2,
		\end{align}
		where the equality $(a)$ comes from Condition 3 in \underline{A1}, and the equality $(b)$ is due to $\mathbf{a}^{\top}\mathbf{b} = \frac{1}{2} \left(\|\mathbf{a}\|^2 + \|\mathbf{b}\|^2 - \|\mathbf{a}-\mathbf{b}\|^2 \right)$.
	\end{proof}
	
	\noindent
	\rm\textbf{Lemma 3.} \textit{Under Condition 3 and 4 in \underline{A1}, the squared norm of the average mini-batch gradients is bounded by}
	\begin{equation}
		\mathbb{E}\|\mathcal{G}_k\|^2 \le \left(\frac{\beta}{m^2} + \frac{1}{m}\right) \sum_{i=1}^{m} \left\|\nabla F(\theta_{k}^{(i)})\right\|^2 + \frac{\sigma^2}{m}.
	\end{equation}
	\begin{proof}
		According to the definition of $\mathcal{G}_k$, we have
		\begin{align}
			 & \mathbb{E}\|\mathcal{G}_k\|^2 = \mathbb{E}\left\|\mathcal{G}_k - \mathbb{E}\left[\mathcal{G}_k\right]\right\|^2 + \left\| \mathbb{E}\left[\mathcal{G}_k\right] \right\|^2 \notag                        \\
			 & = \mathbb{E}\|\mathcal{G}_k - \mathcal{H}_k\|^2 + \|\mathcal{H}_k\|^2 \notag                                                                                                                            \\
			 & \stackrel{(a)}{\le} \frac{\beta}{m^2} \sum_{i=1}^{m} \left\|\nabla F(\theta_{k}^{(i)})\right\|^2 + \frac{\sigma^2}{m} + \left\| \frac{1}{m} \sum_{i=1}^{m} \nabla F(\theta_{k}^{(i)}) \right\|^2 \notag \\
			 & \stackrel{(b)}{\le} \frac{\beta}{m^2} \sum_{i=1}^{m} \left\|\nabla F(\theta_{k}^{(i)})\right\|^2 + \frac{\sigma^2}{m} + \frac{1}{m} \sum_{i=1}^{m} \left\|\nabla F(\theta_{k}^{(i)})\right\|^2 \notag   \\
			 & = \left(\frac{\beta}{m^2} + \frac{1}{m}\right) \sum_{i=1}^{m} \left\|\nabla F(\theta_{k}^{(i)})\right\|^2 + \frac{\sigma^2}{m}, \notag
		\end{align}
		where the inequality $(a)$ is due to Lemma 1, and the inequality $(b)$ comes from the convexity of the vector norm and Jensen's inequality
		\begin{equation}
			\left\| \sum_{i=1}^{m} \mathbf{a}_i \right\|^2 \le m \sum_{i=1}^{m} \left\| \mathbf{a}_i \right\|^2. \notag
		\end{equation}
	\end{proof}
	
	\noindent
	\rm\textbf{Lemma 4.} \textit{Under \underline{A1}, if the total number of iterations $K$ is large enough, and the learning rate $\eta$ satisfies}
	\begin{equation}
		\eta L\left(\frac{\beta}{m} + 1\right) - 1 \le 0,
	\end{equation}
	\textit{then the expected gradient norm after $K$ iterations is bounded by}
	\begin{align}
		 & \mathbb{E}\left[\frac{1}{K} \sum_{k=0}^{K-1} \left\| \nabla F(\bar{\theta}_k) \right\|^2\right] \le \frac{2\left
		[F(\bar{\theta}_0) - F_{\mathrm{inf}}\right]}{\eta K} + \frac{\eta L \sigma^2}{m} \nonumber                                                                                   \\
		 & + \underbrace{\frac{L^2}{mK} \sum_{k=0}^{K-1} \sum_{i=1}^{m} \mathbb{E} \left\| \bar{\theta}_k - \theta_{k}^{(i)} \right\|^2}_{(a)} \label{eq:basic-result-gradient-norm}.
	\end{align}
	\begin{proof}
		According to the Lipschitz continuous gradient assumption, we obtain
		\begin{align}
			 & \mathbb{E} \left[ F(\bar{\theta}_{k+1}) \right] - \mathbb{E} \left[ F(\bar{\theta}_{k}) \right] \notag                                                                                                             \\
			 & \le \mathbb{E}\left[ \left\langle \nabla F(\bar{\theta}_{k}), \bar{\theta}_{k+1} - \bar{\theta}_{k} \right\rangle \right] + \frac{L}{2} \mathbb{E} \left\| \bar{\theta}_{k+1} - \bar{\theta}_{k} \right\|^2 \notag \\
			 & \stackrel{(a)}{=} -\eta \mathbb{E}\left[ \left\langle \nabla F(\bar{\theta}_{k}), \mathcal{G}_k \right\rangle \right] + \frac{\eta^2L}{2} \mathbb{E}\left\| \mathcal{G}_k \right\|^2 \notag                        \\
			 & \stackrel{(b)}{\le} -\frac{\eta}{2} \left\| \nabla F(\bar{\theta}_k) \right\|^2 - \frac{\eta}{2m} \sum_{i=1}^{m} \left\|\nabla F(\theta_{k}^{(i)})\right\|^2 \notag                                                \\
			 & + \frac{\eta}{2m} \sum_{i=1}^{m} \left\| \nabla F(\bar{\theta}_k) - \nabla F(\theta_{k}^{(i)}) \right\|^2 \label{eq:expected_F_function_error}                                                                     \\
			 & + \frac{\eta^2L}{2}\left(\frac{\beta}{m^2} + \frac{1}{m}\right) \sum_{i=1}^{m} \left\|\nabla F(\theta_{k}^{(i)})\right\|^2 + \frac{\eta^2\sigma^2L}{2m},\notag
		\end{align}
		where the equality $(a)$ comes from $\bar{\theta}_{k+1} = \bar{\theta}_{k} - \eta \mathcal{G}_k$, and the inequality $(b)$ is based on Lemma 2 and 3. We apply Condition 1 in \underline{A1} to \eqref{eq:expected_F_function_error} and rearrange the expression to get
		\begin{align}
			 & \left\| \nabla F(\bar{\theta}_k) \right\|^2 \le \frac{2\left[ \mathbb{E}\left[F(\bar{\theta}_k)\right] - \mathbb{E}\left[F(\bar{\theta}_{k+1})\right] \right]}{\eta} + \frac{\eta L \sigma^2}{m} \notag \\
			 & + \left[\eta L\left(\frac{\beta}{m^2} + \frac{1}{m}\right) - \frac{1}{m}\right] \sum_{i=1}^{m} \left\|\nabla F(\theta_{k}^{(i)})\right\|^2 \notag                                                       \\
			 & + \frac{L^2}{m} \sum_{i=1}^{m} \left\| \bar{\theta}_k - \theta_{k}^{(i)} \right\|^2. \notag
		\end{align}
		By superposing and averaging the expression over all iterations $K$, and based on Condition 2 in \underline{A1}, we obtain
		\begin{align}
			 & \mathbb{E}\left[\frac{1}{K} \sum_{k=0}^{K-1} \left\| \nabla F(\bar{\theta}_k) \right\|^2\right] \le \frac{2\left[F(\bar{\theta}_0) - F_{\mathrm{inf}}\right]}{\eta K} + \frac{\eta L \sigma^2}{m} \notag \\
			 & + \left[\eta L\left(\frac{\beta}{m} + 1\right) - 1\right] \frac{1}{mK} \sum_{k=0}^{K-1} \sum_{i=1}^{m} \mathbb{E} \left\| \nabla F(\theta_{k}^{(i)}) \right\|^2 \label{eq:lemma4_pre_final_result}       \\
			 & + \frac{L^2}{mK} \sum_{k=0}^{K-1} \sum_{i=1}^{m} \mathbb{E} \left\| \bar{\theta}_k - \theta_{k}^{(i)} \right\|^2.\notag
		\end{align}
		According to \eqref{eq:lemma4_pre_final_result}, if $\eta L\left(\frac{\beta}{m} + 1\right) - 1 \le 0$, then
		\begin{align}
			 & \mathbb{E}\left[\frac{1}{K} \sum_{k=0}^{K-1} \left\| \nabla F(\bar{\theta}_k) \right\|^2\right] \le \frac{2\left
			[F(\bar{\theta}_0) - F_{\mathrm{inf}}\right]}{\eta K} + \frac{\eta L \sigma^2}{m} \notag                                    \\
			 & + \frac{L^2}{mK} \sum_{k=0}^{K-1} \sum_{i=1}^{m} \mathbb{E} \left\| \bar{\theta}_k - \theta_{k}^{(i)} \right\|^2. \notag
		\end{align}
	\end{proof}
	
	\section{Proof of the result in Section V-B (i.e., (14) and (15))}
	According to the term $(a)$ of \eqref{eq:basic-result-gradient-norm} in Lemma 4, the error in $\sum_{k=0}^{K-1} \sum_{i=1}^{m} \mathbb{E} \left\| \bar{\theta}_k - \theta_{k}^{(i)} \right\|^2$ is due to the discrepancy between different agents. Here, we provide the bound for it. We define the $d \times m$ - dimensional matrix $G_y := \left[ g(\theta_{y}^{(1)}),\ g(\theta_{y}^{(2)}),\ g(\theta_{y}^{(3)}),\ ...,\ g(\theta_{y}^{(m)}) \right]$, and $Y_j := \sum_{y=t_0}^{t_0 + j - 1} G_y$. Here, $d$ denotes the dimension of the model's parameters, and $j=k-t_0$.  
	\begin{proof}
		Since the update rule of the model's parameters in a period can be expressed by
		\begin{equation}
			\label{eq:theta_k_i}
			\theta_{k}^{(i)} = \bar{\theta}_{t_0} - \eta \sum_{y=t_0}^{t_0 + j - 1} g(\theta_{y}^{(i)});
		\end{equation}
		\begin{equation}
			\label{eq:theta_k_bar}
			\bar{\theta}_k = \bar{\theta}_{t_0} - \eta \frac{1}{m} \sum_{i=1}^{m} \sum_{y=t_0}^{t_0 + j - 1} g(\theta_{y}^{(i)}).
		\end{equation}
		By substituting \eqref{eq:theta_k_i} and \eqref{eq:theta_k_bar} into $\sum_{i=1}^{m} \mathbb{E} \left\| \bar{\theta}_k - \theta_{k}^{(i)} \right\|^2$, we have
		\begin{align}
			 & \sum_{i=1}^{m} \mathbb{E} \left\| \bar{\theta}_k - \theta_{k}^{(i)} \right\|^2 \notag                                                                                                    \\
			 & = \eta^2  \sum_{i=1}^{m} \mathbb{E} \left\| \sum_{y=t_0}^{t_0 + j - 1} g(\theta_{y}^{(i)}) - \sum_{y=t_0}^{t_0 + j - 1} \frac{1}{m} \sum_{i=1}^{m} g(\theta_{y}^{(i)}) \right\|^2 \notag \\
			 & = \eta^2 \mathbb{E} \left\| \sum_{y=t_0}^{t_0 + j - 1} G_{y} - \sum_{y=t_0}^{t_0 + j - 1} G_{y} \mathbf{J} \right\|_{\mathrm{F}}^2 \notag                                                \\
			 & = \eta^2 \mathbb{E} \left\| Y_j - Y_j \mathbf{J} \right\|_{\mathrm{F}}^2 \notag                                                                                                          \\
			 & = \eta^2 \mathbb{E} \left\| Y_j(\mathbf{I} - \mathbf{J}) \right\|_{\mathrm{F}}^2 \notag                                                                                                  \\
			 & \stackrel{(a)}{\le} \eta^2 \mathbb{E} \left\| Y_j \right\|_{\mathrm{F}}^2 \ \left\| \mathbf{I} - \mathbf{J} \right\|_{\mathrm{op}}^2 \notag                                              \\
			 & = \eta^2 \mathbb{E} \left\| Y_j \right\|_{\mathrm{F}}^2, \label{eq:expected_theta_error}
		\end{align}
		where the inequality $(a)$ comes from \eqref{eq:operator_frobenius_norm_inequality}. Based on \eqref{eq:expected_theta_error}, we obtain
		\begin{align}
			 & \sum_{i=1}^{m} \mathbb{E} \left\| \bar{\theta}_k - \theta_{k}^{(i)} \right\|^2 \le \eta^2 \sum_{i=1}^{m} \mathbb{E} \left\| \sum_{y=t_0}^{t_0 + j - 1} g(\theta_{y}^{(i)}) \right\|^2 \notag                  \\
			 & = \eta^2 \sum_{i=1}^{m} \mathbb{E} \left\| \sum_{y=t_0}^{t_0 + j - 1} \left(g(\theta_{y}^{(i)}) - \nabla F(\theta_{y}^{(i)})\right) + \sum_{y=t_0}^{t_0 + j - 1} \nabla F(\theta_{y}^{(i)}) \right\|^2 \notag \\
			 & \stackrel{(a)}{\le} 2\eta^2 \sum_{i=1}^{m} \mathbb{E} \left\| \sum_{y=t_0}^{t_0 + j - 1} \left(g(\theta_{y}^{(i)}) - \nabla F(\theta_{y}^{(i)})\right) \right\|^2 \notag                                      \\
			 & + 2\eta^2 \sum_{i=1}^{m} \mathbb{E} \left\| \sum_{y=t_0}^{t_0 + j - 1} \nabla F(\theta_{y}^{(i)}) \right\|^2 \notag                                                                                           \\
			 & \stackrel{(b)}{\le} 2\eta^2 \sum_{i=1}^{m} \Bigg[\sum_{y=t_0}^{t_0 + j - 1} \mathbb{E} \left\| g(\theta_{y}^{(i)}) - \nabla F(\theta_{y}^{(i)}) \right\|^2 \notag                                             \\
			 & + \sum_{y \ne q} \mathbb{E} \left\langle g(\theta_{y}^{(i)}) - \nabla F(\theta_{y}^{(i)}), g(\theta_{q}^{(i)}) - \nabla F(\theta_{q}^{(i)}) \right\rangle  \Bigg] \notag                                      \\
			 & + 2\eta^2 j \sum_{i=1}^{m} \sum_{y=t_0}^{t_0 + j - 1} \mathbb{E} \left\| \nabla F(\theta_{y}^{(i)}) \right\|^2 \notag                                                                                         \\
			 & \stackrel{(c)}{\le} 2\eta^2 \sum_{i=1}^{m} \sum_{y=t_0}^{t_0 + j - 1} \left(\beta \left\| \nabla F(\theta_{y}^{(i)}) \right\|^2 + \sigma^2 \right) \notag                                                     \\
			 & + 2\eta^2 j \sum_{i=1}^{m} \sum_{y=t_0}^{t_0 + j - 1} \left\| \nabla F(\theta_{y}^{(i)}) \right\|^2, \notag
		\end{align}
		where the inequality $(a)$ and $(b)$ come from Jensen's inequality, and according to Condition 3 in \underline{A1}, the term $\sum_{y \ne q} \mathbb{E} \left\langle g(\theta_{y}^{(i)}) - \nabla F(\theta_{y}^{(i)}), g(\theta_{q}^{(i)}) - \nabla F(\theta_{q}^{(i)}) \right\rangle$ turns to be $0$. The inequality $(c)$ comes from Condition 4 in \underline{A1}. By superposing the expression over a period of $\tau$, we get
		\begin{align}
			 & \sum_{j=1}^{\tau} \sum_{i=1}^{m} \mathbb{E} \left\| \bar{\theta}_k - \theta_{k}^{(i)} \right\|^2 \le \eta^2 \sigma^2 m \tau(\tau + 1) \notag                  \\
			 & + \left[2\eta^2 \tau \beta + \eta^2 \tau (\tau + 1)\right] \sum_{i=1}^{m} \sum_{y=t_0}^{t_0 + \tau - 1} \left\| \nabla F(\theta_{y}^{(i)}) \right\|^2. \notag
		\end{align}
		By further superposing the expression over all iterations, we get
		\begin{align}
			 & \sum_{t_0=0}^{K - \tau} \sum_{j=1}^{\tau} \sum_{i=1}^{m} \mathbb{E} \left\| \bar{\theta}_k - \theta_{k}^{(i)} \right\|^2 \le K\eta^2 \sigma^2 m (\tau + 1) \notag \\
			 & + \left[2\eta^2 \tau \beta + \eta^2 \tau (\tau + 1)\right] \sum_{k=0}^{K-1} \sum_{i=1}^{m} \left\| \nabla F(\theta_{k}^{(i)}) \right\|^2. \notag
		\end{align}
		Note that $k = t_0 + j$ and $t_0 = z\tau$, where $z \in \mathbb{N}$. Finally, we get
		\begin{align}
			 & \frac{L^2}{mK} \sum_{k=0}^{K-1} \sum_{i=1}^{m} \mathbb{E} \left\| \bar{\theta}_k - \theta_{k}^{(i)} \right\|^2 \le \eta^2 \sigma^2 L^2 (\tau + 1) \notag               \\
			 & +  \left[2\eta^2 L^2 \tau \beta + \eta^2 L^2 \tau (\tau + 1)\right] \frac{1}{mK} \sum_{k=0}^{K-1} \sum_{i=1}^{m} \left\| \nabla F(\theta_{k}^{(i)}) \right\|^2. \notag
		\end{align}
		By substituting the above inequality into Lemma 4, we obtain
		\begin{align}
			 & \mathbb{E}\left[\frac{1}{K} \sum_{k=0}^{K-1} \left\| \nabla F(\bar{\theta}_k) \right\|^2\right] \le \frac{2\left[F(\bar{\theta}_0) - F_{\mathrm{inf}}\right]}{\eta K} + \frac{\eta L \sigma^2}{m} \notag \\
			 & +  \eta^2 \sigma^2 L^2(\tau + 1) + \Big[2\eta^2L^2\tau\beta + \eta^2L^2\tau(\tau + 1) \notag                                                                                                             \\
			 & + \eta L \left(\frac{\beta}{m} + 1\right) - 1 \Big] \frac{1}{mK} \sum_{k=0}^{K-1} \sum_{i=1}^{m} \mathbb{E} \left\| \nabla F(\theta_{k}^{(i)}) \right\|^2. \notag
		\end{align}
		If $2\eta^2L^2\tau\beta + \eta^2L^2\tau(\tau + 1) + \eta L\left(\frac{\beta}{m} + 1\right) - 1 \le 0$, then we can get the following conclusion
		\begin{align}
			\mathbb{E}\left[\frac{1}{K} \sum_{k=0}^{K-1} \left\| \nabla F(\bar{\theta}_k) \right\|^2\right] & \le \frac{2\left[F(\bar{\theta}_0) - F_{\mathrm{inf}}\right]}{\eta K} + \frac{\eta L \sigma^2}{m} \notag \\
			                                                                                                & +  \eta^2 \sigma^2 L^2(\tau + 1). \notag
		\end{align}
	\end{proof}
	
	\section{Proof of the result in Section V-C (i.e., (17))}
	Since the conclusion is also based on \underline{A1}, we can find that the additional \underline{A2} only affects $\sum_{k=0}^{K-1} \sum_{i=1}^{m} \mathbb{E} \left\| \bar{\theta}_k - \theta_{k}^{(i)} \right\|^2$ while keeping the other terms in the conclusion of Lemma 4 unchanged. Therefore, we directly begin our proofs from Lemma 4 to provide the corresponding bound for $\frac{L^2}{mK} \sum_{k=0}^{K-1} \sum_{i=1}^{m} \mathbb{E} \left\| \bar{\theta}_k - \theta_{k}^{(i)} \right\|^2$. In particular, we define the $m$ - dimensional vector $M_y := \Big[\mathrm{I}(\tau_1 > y - t_0),\ \mathrm{I}(\tau_2 > y-t_0),\ \mathrm{I}(\tau_3 > y-t_0),\ ...,\ \mathrm{I}(\tau_m > y-t_0)\Big]$, and the matrix $P_y := M_y \odot G_y$, where $\odot$ denotes the Hadamard product. Moreover, for $k = t_0 + j$, the summation of $P_y$ over a length of $j$ can be expressed by $Q_j := \sum_{y=t_0}^{t_0 + j - 1}P_{y}$. 
	\begin{proof}
		Since the update rule  of the model's parameters in a period is expressed by
		\begin{equation}
			\label{eq:theta_k_i_vpa}
			\theta_{k}^{(i)} = \bar{\theta}_{t_0} - \eta \sum_{y=t_0}^{t_0 + j - 1} \mathrm{I}(\tau_{i}>y-t_0)\ g(\theta_{y}^{(i)});
		\end{equation}
		
		\begin{equation}
			\label{eq:theta_k_bar_vpa}
			\bar{\theta}_k = \bar{\theta}_{t_0} - \eta \frac{1}{m} \sum_{i=1}^{m} \sum_{y=t_0}^{t_0 + j - 1} \mathrm{I}(\tau_{i}>y-t_0)\ g(\theta_{y}^{(i)}).
		\end{equation}
		By substituting \eqref{eq:theta_k_i_vpa} and \eqref{eq:theta_k_bar_vpa} into $\sum_{i=1}^{m} \mathbb{E} \left\| \bar{\theta}_k - \theta_{k}^{(i)} \right\|^2$, we obtain
		\begin{align}
			 & \sum_{i=1}^{m} \mathbb{E} \left\| \bar{\theta}_k - \theta_{k}^{(i)} \right\|^2 = \eta^2  \sum_{i=1}^{m} \mathbb{E} \left\| \sum_{y=t_0}^{t_0 + j - 1} \mathrm{I}(\tau_{i}>y-t_0)\ g(\theta_{y}^{(i)}) \right. \notag \\
			 & \left. - \sum_{y=t_0}^{t_0 + j - 1} \frac{1}{m} \sum_{i=1}^{m} \mathrm{I}(\tau_{i}>y-t_0)\ g(\theta_{y}^{(i)}) \right\|^2 \notag                                                                                     \\
			 & = \eta^2 \mathbb{E} \left\| \sum_{y=t_0}^{t_0 + j - 1} P_{y} - \sum_{y=t_0}^{t_0 + j - 1} P_{y} \mathbf{J} \right\|_{\mathrm{F}}^2 \notag                                                                            \\
			 & = \eta^2 \mathbb{E} \left\| Q_j - Q_j \mathbf{J} \right\|_{\mathrm{F}}^2 \notag                                                                                                                                      \\
			 & = \eta^2 \mathbb{E} \left\| Q_j(\mathbf{I} - \mathbf{J}) \right\|_{\mathrm{F}}^2 \notag                                                                                                                              \\
			 & \le \eta^2 \mathbb{E} \left\| Q_j \right\|_{\mathrm{F}}^2 \ \left\| \mathbf{I} - \mathbf{J} \right\|_{\mathrm{op}}^2 \notag                                                                                          \\
			 & = \eta^2 \mathbb{E} \left\| Q_j \right\|_{\mathrm{F}}^2 \notag                                                                                                                                                       \\
			 & = \eta^2 \sum_{i=1}^{m} \mathbb{E} \left\| \sum_{y=t_0}^{t_0 + j - 1} \mathrm{I}(\tau_{i}>y-t_0)\ g(\theta_{y}^{(i)}) \right\|^2 \notag                                                                              \\
			 & = \eta^2 \sum_{i=1}^{m} \mathbb{E} \left\| \sum_{y=t_0}^{t_0 + j - 1} \mathrm{I}(\tau_{i}>y-t_0)\ \left(g(\theta_{y}^{(i)}) - \nabla F(\theta_{y}^{(i)})\right) \right. \notag                                       \\
			 & \left. + \sum_{y=t_0}^{t_0 + j - 1} \mathrm{I}(\tau_{i}>y-t_0)\ \nabla F(\theta_{y}^{(i)}) \right\|^2 \label{eq:expected_theta_error_vpa}                                                                            \\
			 & \le 2\eta^2 \sum_{i=1}^{m} \mathbb{E} \left\| \sum_{y=t_0}^{t_0 + j - 1} \mathrm{I}(\tau_{i}>y-t_0)\left(g(\theta_{y}^{(i)}) - \nabla F(\theta_{y}^{(i)})\right) \right\|^2 \notag                                   \\
			 & + 2\eta^2 \sum_{i=1}^{m} \mathbb{E} \left\| \sum_{y=t_0}^{t_0 + j - 1} \mathrm{I}(\tau_{i}>y-t_0)\ \nabla F(\theta_{y}^{(i)}) \right\|^2 \notag                                                                      \\
			 & \le 2\eta^2 \sum_{i=1}^{m} \Bigg[\sum_{y=t_0}^{t_0 + j - 1} \mathbb{E} \left\| \mathrm{I}(\tau_{i}>y-t_0)\left( g(\theta_{y}^{(i)}) - \nabla F(\theta_{y}^{(i)})\right) \right\|^2 \notag                            \\
			 & + \sum_{y \ne q} \mathbb{E} \Big\langle \mathrm{I}(\tau_{i}>y-t_0)\ \left(g(\theta_{y}^{(i)}) - \nabla F(\theta_{y}^{(i)})\right), \notag                                                                            \\
			 & \mathrm{I}(\tau_{i}>q-t_0)\ \left(g(\theta_{q}^{(i)}) - \nabla F(\theta_{q}^{(i)})\right) \Big\rangle  \Bigg] \notag                                                                                                 \\
			 & + 2\eta^2 j \sum_{i=1}^{m} \sum_{y=t_0}^{t_0 + j - 1} \mathbb{E} \left\| \mathrm{I}(\tau_{i}>y-t_0)\ \nabla F(\theta_{y}^{(i)}) \right\|^2. \notag
		\end{align}
		Since \underline{A2} does not affect Condition 3 in \underline{A1}, the term $\sum_{y \ne q} \mathbb{E} \Big\langle \mathrm{I}(\tau_{i}>y-t_0)\ \left(g(\theta_{y}^{(i)}) - \nabla F(\theta_{y}^{(i)})\right),\ \mathrm{I}(\tau_{i}>q-t_0)\ \left(g(\theta_{q}^{(i)}) - \nabla F(\theta_{q}^{(i)})\right) \Big\rangle$ turns to be $0$. Moreover, by applying Condition 4 in \underline{A1} to \eqref{eq:expected_theta_error_vpa}, we obtain
		\begin{align}
			 & \sum_{i=1}^{m} \mathbb{E} \left\| \bar{\theta}_k - \theta_{k}^{(i)} \right\|^2 \notag                                                                                                        \\
			 & \le \underbrace{2\eta^2 \sum_{i=1}^{m} \sum_{y=t_0}^{t_0 + j - 1} \mathrm{I}(\tau_{i}>y-t_0)\left[\beta \left\| \nabla F(\theta_{y}^{(i)}) \right\|^2 + \sigma^2 \right]}_{(a)} \nonumber    \\
			 & + \underbrace{2\eta^2 j \sum_{i=1}^{m} \sum_{y=t_0}^{t_0 + j - 1} \mathrm{I}(\tau_{i}>y-t_0)\left\| \nabla F(\theta_{y}^{(i)}) \right\|^2}_{(b)}. \label{eq:expected_theta_error_vpa_update}
		\end{align}
		By superposing the expression over a period of $\tau$, we find
		\begin{align}
			 & \sum_{j=1}^{\tau} \sum_{i=1}^{m} \mathbb{E} \left\| \bar{\theta}_k - \theta_{k}^{(i)} \right\|^2 \le 2\eta^2 \sigma^2 \sum_{i=1}^{m}\sum_{j=1}^{\tau} \mathrm{min}\{\tau_{i},\ j\}\ + \notag \\
			 & \left[2\eta^2 \tau \beta + \eta^2 \tau (\tau + 1)\right] \sum_{i=1}^{m} \sum_{y=t_0}^{t_0 + \tau - 1} \mathrm{I}(\tau_{i}>y-t_0) \left\| \nabla F(\theta_{y}^{(i)}) \right\|^2. \notag
		\end{align}
		By further superposing the expression over all iterations, we finally obtain
		\begin{align}
			 & \sum_{t_0=0}^{K - \tau} \sum_{j=1}^{\tau} \sum_{i=1}^{m} \mathbb{E} \left\| \bar{\theta}_k - \theta_{k}^{(i)} \right\|^2 \le \left[2\eta^2 \tau \beta + \eta^2 \tau (\tau + 1)\right] \notag \\
			 & \sum_{i=1}^{m} \sum_{t_0=0}^{K - \tau} \sum_{y=t_0}^{t_0 + \tau - 1} \mathrm{I}(\tau_{i}>y-t_0) \left\| \nabla F(\theta_{y}^{(i)}) \right\|^2 \notag                                         \\
			 & + 2\eta^2 \sigma^2 \sum_{i=1}^{m} \sum_{t_0=0}^{K - \tau} \sum_{j=1}^{\tau} \mathrm{min}\{\tau_{i},\ j\} \notag                                                                              \\
			 & = \left[2\eta^2 \tau \beta + \eta^2 \tau (\tau + 1)\right] \sum_{k=0}^{K-1} \sum_{i=1}^{m} \mathrm{I}(\tau_{i} > k\ \mathrm{mod}\ \tau) \notag                                               \\
			 & \left\| \nabla F(\theta_{k}^{(i)}) \right\|^2 + 2\eta^2 \sigma^2 \frac{K}{\tau} \sum_{i=1}^{m} \sum_{j=1}^{\tau} \mathrm{min}\{\tau_{i},\ j\}. \notag
		\end{align}
		Therefore, the term $(a)$ of \eqref{eq:basic-result-gradient-norm} in Lemma 4 can be bounded by
		\begin{align}
			 & \frac{L^2}{mK} \sum_{k=0}^{K-1} \sum_{i=1}^{m} \mathbb{E} \left\| \bar{\theta}_k - \theta_{k}^{(i)} \right\|^2 \le \left[2\eta^2 L^2 \tau \beta + \eta^2 L^2 \tau (\tau + 1)\right] \notag \\
			 & \frac{1}{mK} \sum_{k=0}^{K-1} \sum_{i=1}^{m} \mathrm{I}(\tau_{i} > k\ \mathrm{mod}\ \tau) \left\| \nabla F(\theta_{k}^{(i)}) \right\|^2 \notag                                             \\
			 & + \frac{1}{m\tau} 2\eta^2 L^2 \sigma^2 \sum_{i=1}^{m} \sum_{j=1}^{\tau} \mathrm{min}\{\tau_{i},\ j\} \notag                                                                                \\
			 & \le \left[2\eta^2 L^2 \tau \beta + \eta^2 L^2 \tau (\tau + 1)\right] \frac{1}{mK} \sum_{k=0}^{K-1} \sum_{i=1}^{m} \left\| \nabla F(\theta_{k}^{(i)}) \right\|^2 \notag                     \\
			 & + \underbrace{\frac{\eta^2 L^2 \sigma^2}{\tau} \frac{1}{m}\sum_{i=1}^{m} \left(\tau_{i} + 2\tau \tau_{i} - \tau_{i}^2 \right)}_{(a)} \label{eq:result-gradient-norm_vpa}. 
		\end{align}
		Based on Condition 4 and 5 in \underline{A2}, we can estimate the expectation of the term $(a)$ in \eqref{eq:result-gradient-norm_vpa} when the total number of iterations $K$ is large enough. We have
		\begin{align}
			 & \frac{L^2}{mK} \sum_{k=0}^{K-1} \sum_{i=1}^{m} \mathbb{E} \left\| \bar{\theta}_k - \theta_{k}^{(i)} \right\|^2 \notag                                                  \\
			 & \le \left[2\eta^2 L^2 \tau \beta + \eta^2 L^2 \tau (\tau + 1)\right] \frac{1}{mK} \sum_{k=0}^{K-1} \sum_{i=1}^{m} \left\| \nabla F(\theta_{k}^{(i)}) \right\|^2 \notag \\
			 & + \frac{\eta^2 \sigma^2 L^2}{\tau} \left[-\nu^2 +(2\tau + 1)\nu - w^2 \right]. \notag
		\end{align}
		By substituting the above inequality into the term $(a)$ of \eqref{eq:basic-result-gradient-norm} in Lemma 4, we prove the conclusion.
	\end{proof}
	
	\section{Proof of Theorem 1}
	According to \underline{A3}, we define the $d \times m$ - dimensional matrix $G_y^{'} := \Big[D(y)g(\theta_{y}^{(1)}),\ D(y)g(\theta_{y}^{(2)}),\ D(y)g(\theta_{y}^{(3)}),\ ...,\ D(y)g(\theta_{y}^{(m)})\Big]$, and $P_y^{'} := M_y \odot G_y^{'}$. Then, the summation of $P_y^{'}$ over a length of $j$ can be expressed by $Q_j^{'} := \sum_{y=t_0}^{t_0 + j - 1}P_{y}^{'}$. Since the conclusion in Theorem 1 is based on \underline{A1}, we begin our proofs directly from providing the bound of $\frac{L^2}{mK} \sum_{k=0}^{K-1} \sum_{i=1}^{m} \mathbb{E} \left\| \bar{\theta}_k - \theta_{k}^{(i)} \right\|^2$ in Lemma 4.
	\begin{proof}
		Since the update rule  of the model's parameters in a period can be expressed by
		\begin{equation}
			\label{eq:theta_k_i_decay}
			\theta_{k}^{(i)} = \bar{\theta}_{t_0} - \eta \sum_{y=t_0}^{t_0 + j - 1} \mathrm{I}(\tau_{i}>y-t_0)\ D(y)\ g(\theta_{y}^{(i)});
		\end{equation}
		\begin{equation}
			\label{eq:theta_k_bar_decay}
			\bar{\theta}_k = \bar{\theta}_{t_0} - \eta \frac{1}{m} \sum_{i=1}^{m} \sum_{y=t_0}^{t_0 + j - 1} \mathrm{I}(\tau_{i}>y-t_0)\ D(y)\ g(\theta_{y}^{(i)}).
		\end{equation}
		By substituting \eqref{eq:theta_k_i_decay} and \eqref{eq:theta_k_bar_decay} into $\sum_{i=1}^{m} \mathbb{E} \left\| \bar{\theta}_k - \theta_{k}^{(i)} \right\|^2$, we obtain
		\begin{align}
			 & \sum_{i=1}^{m} \mathbb{E} \left\| \bar{\theta}_k - \theta_{k}^{(i)} \right\|^2 \notag                                                                                            \\
			 & = \eta^2  \sum_{i=1}^{m} \mathbb{E} \left\| \sum_{y=t_0}^{t_0 + j - 1} \mathrm{I}(\tau_{i}>y-t_0)\ D(y)\ g(\theta_{y}^{(i)}) \right. \notag                                      \\
			 & \left. - \sum_{y=t_0}^{t_0 + j - 1} \frac{1}{m} \sum_{i=1}^{m} \mathrm{I}(\tau_{i}>y-t_0)\ D(y)\ g(\theta_{y}^{(i)}) \right\|^2 \notag                                           \\
			 & = \eta^2 \mathbb{E} \left\| \sum_{y=t_0}^{t_0 + j - 1} P_{y}^{'} - \sum_{y=t_0}^{t_0 + j - 1} P_{y}^{'} \mathbf{J} \right\|_{\mathrm{F}}^2 \notag                                \\
			 & = \eta^2 \mathbb{E} \left\| Q_j^{'} - Q_j^{'} \mathbf{J} \right\|_{\mathrm{F}}^2 \notag                                                                                          \\
			 & = \eta^2 \mathbb{E} \left\| Q_j^{'}(\mathbf{I} - \mathbf{J}) \right\|_{\mathrm{F}}^2 \notag                                                                                      \\
			 & \le \eta^2 \mathbb{E} \left\| Q_j^{'} \right\|_{\mathrm{F}}^2 \ \left\| \mathbf{I} - \mathbf{J} \right\|_{\mathrm{op}}^2 \notag                                                  \\
			 & = \eta^2 \mathbb{E} \left\| Q_j^{'} \right\|_{\mathrm{F}}^2 \notag                                                                                                               \\
			 & = \eta^2 \sum_{i=1}^{m} \mathbb{E} \left\| \sum_{y=t_0}^{t_0 + j - 1} \mathrm{I}(\tau_{i}>y-t_0)\ D(y)\ g(\theta_{y}^{(i)}) \right\|^2 \notag                                    \\
			 & = \eta^2 \sum_{i=1}^{m} \mathbb{E} \left\| \sum_{y=t_0}^{t_0 + j - 1} \mathrm{I}(\tau_{i}>y-t_0)D(y)\left(g(\theta_{y}^{(i)}) - \nabla F(\theta_{y}^{(i)})\right) \right. \notag \\
			 & \left. + \sum_{y=t_0}^{t_0 + j - 1} \mathrm{I}(\tau_{i}>y-t_0)\ D(y)\ \nabla F(\theta_{y}^{(i)}) \right\|^2 \notag                                                               \\
			 & \le 2\eta^2 \sum_{i=1}^{m} \mathbb{E} \Bigg\| \sum_{y=t_0}^{t_0 + j - 1} \mathrm{I}(\tau_{i}>y-t_0)\ D(y) \notag                                                                 \\
			 & \left(g(\theta_{y}^{(i)}) - \nabla F(\theta_{y}^{(i)})\right) \Bigg\|^2 \notag                                                                                                   \\
			 & + 2\eta^2 \sum_{i=1}^{m} \mathbb{E} \left\| \sum_{y=t_0}^{t_0 + j - 1} \mathrm{I}(\tau_{i}>y-t_0)\ D(y)\ \nabla F(\theta_{y}^{(i)}) \right\|^2 \notag                            \\
			 & \le 2\eta^2 \sum_{i=1}^{m} \Bigg[\sum_{y=t_0}^{t_0 + j - 1} \mathbb{E} \Big\| \mathrm{I}(\tau_{i}>y-t_0) D(y) \notag                                                             \\
			 & \left( g(\theta_{y}^{(i)}) - \nabla F(\theta_{y}^{(i)})\right) \Big\|^2 \label{eq:expected_theta_error_decay}                                                                    \\
			 & + \sum_{y \ne q} \mathbb{E} \Big\langle \mathrm{I}(\tau_{i}>y-t_0)D(y)\left(g(\theta_{y}^{(i)}) - \nabla F(\theta_{y}^{(i)})\right), \notag                                      \\
			 & \mathrm{I}(\tau_{i}>q-t_0)\ D(q)\ \left(g(\theta_{q}^{(i)}) - \nabla F(\theta_{q}^{(i)})\right) \Big\rangle  \Bigg] \notag                                                       \\
			 & + 2\eta^2 j \sum_{i=1}^{m} \sum_{y=t_0}^{t_0 + j - 1} \mathbb{E} \left\| \mathrm{I}(\tau_{i}>y-t_0)\ D(y)\ \nabla F(\theta_{y}^{(i)}) \right\|^2.	\notag
		\end{align}
		Since $D(y)$ is independent from the mini-batch gradients and according to Condition 3 in \underline{A1}, we can find that the  term $\sum_{y \ne q} \mathbb{E} \Big\langle \mathrm{I}(\tau_{i}>y-t_0)D(y)\left(g(\theta_{y}^{(i)}) - \nabla F(\theta_{y}^{(i)})\right),\ \mathrm{I}(\tau_{i}>q-t_0)\ D(q)\ \left(g(\theta_{q}^{(i)}) - \nabla F(\theta_{q}^{(i)})\right) \Big\rangle$ turns to be $0$. By further applying Condition 4 in \underline{A1} to \eqref{eq:expected_theta_error_decay}, we can find
		\begin{align}
			 & \sum_{i=1}^{m} \mathbb{E} \left\| \bar{\theta}_k - \theta_{k}^{(i)} \right\|^2 \notag                                                                                                                      \\
			 & \le 2\eta^2 \sum_{i=1}^{m} \sum_{y=t_0}^{t_0 + j - 1} \mathrm{I}(\tau_{i}>y-t_0) D^2(y)\left[\beta \left\| \nabla F(\theta_{y}^{(i)}) \right\|^2 + \sigma^2 \right] \notag                                 \\
			 & + 2\eta^2 j \sum_{i=1}^{m} \sum_{y=t_0}^{t_0 + j - 1} \mathrm{I}(\tau_{i}>y-t_0)D^2(y)\left\| \nabla F(\theta_{y}^{(i)}) \right\|^2 \notag                                                                 \\
			 & \stackrel{(a)}{\le} \underbrace{2\eta^2 \sum_{i=1}^{m} \sum_{y=t_0}^{t_0 + j - 1} \mathrm{I}(\tau_{i}>y-t_0)\left[\beta \left\| \nabla F(\theta_{y}^{(i)}) \right\|^2 + \sigma^2 \right] }_{(a)} \nonumber \\
			 & + \underbrace{2\eta^2 j \sum_{i=1}^{m} \sum_{y=t_0}^{t_0 + j - 1} \mathrm{I}(\tau_{i}>y-t_0)\left\| \nabla F(\theta_{y}^{(i)}) \right\|^2}_{(b)},\label{eq:expected_theta_error_decay_update}
		\end{align}
		where the inequality $(a)$ is due to $D^2(y) \le 1$. Notice that the terms $(a)$ and $(b)$ in \eqref{eq:expected_theta_error_decay_update} are the same as those in \eqref{eq:expected_theta_error_vpa_update}, which comprise the bound of the variation-aware periodic averaging method. Therefore, the model's error convergence bound under the decay-based method is decreased, and Theorem 1 is proved.
	\end{proof}
	
	\section{Proof of Corollary 1}
	Since the conclusion is given as an example of the decay-based method, we can directly begin our proofs on the basis of the proof of Theorem 1. We define the summation of $D^2(y)$ over a length of $j$ as $Z(j) := \sum_{y=t_0}^{t_0 + j - 1} D^2(y)$, then $Z(j) \le j$. According to the definition of $D(y)$, we can get $Z(j) = \frac{1-\lambda^j}{1-\lambda}$.
	\begin{proof}
		According to the proof of Theorem 1, we obtain
		\begin{align}
			 & \sum_{i=1}^{m} \mathbb{E} \left\| \bar{\theta}_k - \theta_{k}^{(i)} \right\|^2 \notag                                                                                      \\
			 & \le 2\eta^2 \sum_{i=1}^{m} \sum_{y=t_0}^{t_0 + j - 1} \mathrm{I}(\tau_{i}>y-t_0) D^2(y)\left[\beta \left\| \nabla F(\theta_{y}^{(i)}) \right\|^2 + \sigma^2 \right] \notag \\
			 & + 2\eta^2 j \sum_{i=1}^{m} \sum_{y=t_0}^{t_0 + j - 1} \mathrm{I}(\tau_{i}>y-t_0)D^2(y)\left\| \nabla F(\theta_{y}^{(i)}) \right\|^2. \notag
		\end{align}
		By superposing the expression over a period of $\tau$, we find
		\begin{align}
			 & \sum_{j=1}^{\tau} \sum_{i=1}^{m} \mathbb{E} \left\| \bar{\theta}_k - \theta_{k}^{(i)} \right\|^2 \le \left[2\eta^2 \tau \beta + \eta^2 \tau (\tau + 1)\right] \notag \\
			 & \sum_{i=1}^{m} \sum_{y=t_0}^{t_0 + \tau - 1}\ \mathrm{I}(\tau_{i}>y-t_0)\ D^2(y)\left\| \nabla F(\theta_{y}^{(i)}) \right\|^2 \notag                                 \\
			 & +2\eta^2 \sigma^2 \sum_{i=1}^{m}\sum_{j=1}^{\tau} \sum_{y=t_0}^{t_0 + j - 1} \mathrm{I}(\tau_{i}>y-t_0)\ D^2(y) \notag                                               \\
			 & = \left[2\eta^2 \tau \beta + \eta^2 \tau (\tau + 1)\right] \sum_{i=1}^{m} \sum_{y=t_0}^{t_0 + \tau - 1} \mathrm{I}(\tau_{i}>y-t_0) \notag                            \\ &D^2(y)\left\| \nabla F(\theta_{y}^{(i)}) \right\|^2 +2\eta^2 \sigma^2 \sum_{i=1}^{m}\sum_{j=1}^{\tau} \mathrm{min}\left\{Z(\tau_{i}),\ Z(j)\right\}. \notag
		\end{align}
		By further superposing the expression over all iterations, we finally obtain
		\begin{align}
			 & \sum_{t_0=0}^{K - \tau} \sum_{j=1}^{\tau} \sum_{i=1}^{m} \mathbb{E} \left\| \bar{\theta}_k - \theta_{k}^{(i)} \right\|^2 \le \left[2\eta^2 \tau \beta + \eta^2 \tau (\tau + 1)\right] \notag \\
			 & \sum_{i=1}^{m} \sum_{t_0=0}^{K-\tau}\sum_{y=t_0}^{t_0 + \tau - 1} \mathrm{I}(\tau_{i}>y-t_0) D^2(y) \left\| \nabla F(\theta_{y}^{(i)}) \right\|^2 \notag                                     \\
			 & + 2\eta^2 \sigma^2 \frac{K}{\tau} \sum_{i=1}^{m} \sum_{j=1}^{\tau} \mathrm{min}\left\{Z(\tau_{i}),\ Z(j)\right\}. \notag
		\end{align}
		Therefore, the bound of the term $(a)$ of \eqref{eq:basic-result-gradient-norm} in Lemma 4 can be expressed by
		\begin{align}
			 & \frac{L^2}{mK} \sum_{k=0}^{K-1} \sum_{i=1}^{m} \mathbb{E} \left\| \bar{\theta}_k - \theta_{k}^{(i)} \right\|^2 \notag                                                                                                                   \\
			 & \le \left[2\eta^2 L^2 \tau \beta + \eta^2 L^2 \tau (\tau + 1)\right] \frac{1}{mK} \sum_{k=0}^{K-1} \sum_{i=1}^{m} \left\| \nabla F(\theta_{k}^{(i)}) \right\|^2 \notag                                                                  \\
			 & + \frac{2\eta^2 L^2 \sigma^2}{m\tau} \sum_{i=1}^{m} \sum_{j=1}^{\tau} \mathrm{min}\left\{Z(\tau_{i}),\ Z(j)\right\} \notag                                                                                                              \\
			 & = \left[2\eta^2 L^2 \tau \beta + \eta^2 L^2 \tau (\tau + 1)\right] \frac{1}{mK} \sum_{k=0}^{K-1} \sum_{i=1}^{m} \left\| \nabla F(\theta_{k}^{(i)}) \right\|^2 \notag                                                                    \\
			 & + \frac{2\eta^2 L^2 \sigma^2}{m\tau} \sum_{i=1}^{m} \left[\sum_{j=1}^{\tau_{i}}Z(j) + (\tau - \tau_{i})Z(\tau_{i})\right] \label{eq:result-gradient-norm_c1}                                                                            \\
			 & = \left[2\eta^2 L^2 \tau \beta + \eta^2 L^2 \tau (\tau + 1)\right] \frac{1}{mK} \sum_{k=0}^{K-1} \sum_{i=1}^{m} \left\| \nabla F(\theta_{k}^{(i)}) \right\|^2 + \notag                                                                  \\
			 & \frac{2\eta^2 L^2 \sigma^2}{\tau} \underbrace{\frac{1}{m} \sum_{i=1}^{m} \left[\frac{(1-\lambda)\tau_{i} - \lambda(1-\lambda^{\tau_{i}})}{(1-\lambda)^2} + (\tau - \tau_{i})\frac{1-\lambda^{\tau_{i}}}{1-\lambda}\right]}_{(a)}.\notag
		\end{align}
		Next, we estimate the expectation of the term $(a)$ in \eqref{eq:result-gradient-norm_c1} when the total number of iterations $K$ is large enough. Since we suppose that the numbers of local updates from different agents are uniformly distributed across the domain, that is $\frac{1}{m} \sum_{i=1}^{m} \tau_{i} \stackrel{K \rightarrow \infty}{\longrightarrow} \frac{1 + \tau}{2}$, we thus can get $\frac{1}{m} \sum_{i=1}^{m} \lambda^{\tau_{i}} \stackrel{K \rightarrow \infty}{\longrightarrow} \frac{\lambda(1-\lambda^{\tau})}{\tau(1-\lambda)}$, and $\frac{1}{m} \sum_{i=1}^{m} \tau_{i}\lambda^{\tau_{i}} \stackrel{K \rightarrow \infty}{\longrightarrow} \frac{\lambda(1-\lambda^{\tau})}{\tau(1-\lambda)^2} -\frac{\lambda^{\tau+1}}{1-\lambda}$. By substituting these conclusions into \eqref{eq:result-gradient-norm_c1}, we get 
		\begin{align}
			 & \frac{L^2}{mK} \sum_{k=0}^{K-1} \sum_{i=1}^{m} \mathbb{E} \left\| \bar{\theta}_k - \theta_{k}^{(i)} \right\|^2 \notag                                                             \\
			 & \le \left[2\eta^2 L^2 \tau \beta + \eta^2 L^2 \tau (\tau + 1)\right] \frac{1}{mK} \sum_{k=0}^{K-1} \sum_{i=1}^{m} \left\| \nabla F(\theta_{k}^{(i)}) \right\|^2 \notag            \\
			 & + \frac{2\eta^2 L^2 \sigma^2}{\tau} \left[\frac{\tau}{1-\lambda} - \frac{2\lambda}{(1-\lambda)^2} + \frac{\lambda(1+\lambda)(1-\lambda^{\tau})}{\tau(1-\lambda)^3}\right]. \notag
		\end{align}
		By substituting the above inequality into the term $(a)$ of \eqref{eq:basic-result-gradient-norm} in Lemma 4, we get the conclusion.
	\end{proof}
	
	\section{Proof of Theorem 2}
	Since the conclusion in Theorem 2 is based on \underline{A1} and \underline{A2}, we directly begin our proofs from presenting the bound of $\frac{L^2}{mK} \sum_{k=0}^{K-1} \sum_{i=1}^{m} \mathbb{E} \left\| \bar{\theta}_k - \theta_{k}^{(i)} \right\|^2$ in Lemma 4. Similarly, we define the $d \times m$ - dimensional matrix $G_{y, e} := \left[g(\theta_{y}^{(1)}, e),\ g(\theta_{y}^{(2)}, e),\ g(\theta_{y}^{(3)}, e),\ ...,\ g(\theta_{y}^{(m)}, e)\right]$, and $Y_{j, e} := \sum_{y=t_0}^{t_0 + j - 1} G_{y, e}$. According to the consensus algorithm, we can infer that $G_{y, 0} = G_y$, and $G_{y, e+1} = G_{y, e}\mathbf{P}$. Furthermore, we can conclude that $G_{y, e} = G_y\mathbf{P}^e$ and $Y_{j, e} = Y_{j}\mathbf{P}^e$, where $e$ denotes the power exponent. Here, the matrix $\mathbf{P} := \mathbf{I} - \epsilon \mathbf{La}$. For the network of agents, the Laplace matrix $\mathbf{La}$ is determined by
	\begin{equation}
		a_{i,l} = \left\{
		\begin{array}{ll}
			|\Omega_{i}|, & i=l;                       \\
			-1,           & i \ne l, l \in \Omega_{i}; \\
			0,            & \mathrm{otherwise}, 
		\end{array}
		\right.
	\end{equation}
	where $a_{i,l}$ is the element in $\mathbf{La}$. In addition, we define $\mu$ as the eigenvalue of $\mathbf{La}$. Then, according to \underline{A4}, we can obtain that $\mu_{min}(\mathbf{La}) = 0$, and the number of eigenvalues that are equal to $0$ is $1$. 
	\begin{proof}
		Since the update rule  of the model's parameters in a period can be expressed by
		\begin{equation}
			\label{eq:theta_k_i_consensus}
			\theta_{k}^{(i)} = \bar{\theta}_{t_0} - \eta \sum_{y = t_0}^{t_0 + j - 1} g(\theta_{y}^{(i)}, e);
		\end{equation}
		\begin{equation}
			\label{eq:theta_k_bar_consensus}
			\bar{\theta}_{k} = \bar{\theta}_{t_0} - \eta \frac{1}{m} \sum_{i = 1}^{m} \sum_{y = t_0}^{t_0 + j - 1} g(\theta_{y}^{(i)}, e).
		\end{equation}
		By substituting \eqref{eq:theta_k_i_consensus} and \eqref{eq:theta_k_bar_consensus} into $\sum_{i=1}^{m} \mathbb{E} \left\| \bar{\theta}_k - \theta_{k}^{(i)} \right\|^2$, we obtain
		\begin{align}
			 & \sum_{i=1}^{m} \mathbb{E} \left\| \bar{\theta}_k - \theta_{k}^{(i)} \right\|^2 \notag                                                                                                          \\
			 & = \eta^2  \sum_{i=1}^{m} \mathbb{E} \left\| \sum_{y=t_0}^{t_0 + j - 1} g(\theta_{y}^{(i)}, e) - \frac{1}{m} \sum_{i=1}^{m} \sum_{y=t_0}^{t_0 + j - 1} g(\theta_{y}^{(i)}, e) \right\|^2 \notag \\
			 & = \eta^2 \mathbb{E} \left\| \sum_{y=t_0}^{t_0 + j - 1} G_{y,e} - \sum_{y=t_0}^{t_0 + j - 1} G_{y,e} \mathbf{J} \right\|_{\mathrm{F}}^2 \notag                                                  \\
			 & = \eta^2 \mathbb{E} \left\| Y_{j,e} - Y_{j,e} \mathbf{J} \right\|_{\mathrm{F}}^2 \notag                                                                                                        \\
			 & = \eta^2 \mathbb{E} \left\| Y_{j,e}(\mathbf{I} - \mathbf{J}) \right\|_{\mathrm{F}}^2 \notag                                                                                                    \\
			 & = \eta^2 \mathbb{E} \left\| Y_{j}\mathbf{P}^{e}(\mathbf{I} - \mathbf{J}) \right\|_{\mathrm{F}}^2 \notag                                                                                        \\
			 & \le \eta^2 \mathbb{E} \left\| Y_j \right\|_{\mathrm{F}}^2 \ \left\|\mathbf{P}^{e} (\mathbf{I} - \mathbf{J}) \right\|_{\mathrm{op}}^2. \label{eq:result-gradient-norm_consensus}
		\end{align}
		According to the operator norm defined in \eqref{eq:operator_norm}, we find
		\begin{align}
			\left\|\mathbf{P}^{e} (\mathbf{I} - \mathbf{J}) \right\|_{\mathrm{op}}^2 & = \lambda_{\mathrm{max}}\left[(\mathbf{P}^{e}(\mathbf{I} - \mathbf{J}))^{\top}(\mathbf{P}^{e}(\mathbf{I} - \mathbf{J}))\right] \\
			                                                                         & = \lambda_{\mathrm{max}}\left[(\mathbf{I} - \mathbf{J})\mathbf{P}^{2e}(\mathbf{I} - \mathbf{J})\right]. 
		\end{align}
		We define $\zeta$ as the eigenvalue of $\mathbf{P}$. Since $\mathbf{P}$ is a real symmetric matrix, it thus can be decomposed by $\mathbf{P} = \mathbf{Q}\mathbf{\Lambda}_P\mathbf{Q}^{\top}$, where $\mathbf{Q}$ is an orthogonal matrix and $diag(\mathbf{\Lambda}_P)=\{\zeta_1,\ \zeta_2,\ \zeta_3,\ ...,\ \zeta_{m}\}$. Similarly, we also have $\mathbf{I} - \mathbf{J} = \mathbf{Q}\mathbf{\Lambda}_0\mathbf{Q}^{\top}$, and $diag(\mathbf{\Lambda}_0) = \{0,\ 1,\ 1,\ ...,\ 1\}$. Then, we can infer that 
		\begin{equation}
			(\mathbf{I} - \mathbf{J})\mathbf{P}^{2e}(\mathbf{I} - \mathbf{J}) = \mathbf{Q}\mathbf{\Lambda}\mathbf{Q}^{\top}, 
		\end{equation}
		where $diag(\mathbf{\Lambda}) = \{0,\ \zeta_2^{2e},\ \zeta_3^{2e},\ ...,\ \zeta_m^{2e}\}$. Since $\mathbf{P} = \mathbf{I} - \epsilon \mathbf{La}$, the corresponding eigenvalue $\zeta$ and $\mu$ satisfy
		\begin{equation}
			\zeta_i = 1 - \epsilon \mu_i,
		\end{equation}
		where $i$ denotes the index of eigenvalue. Since $\mathbf{La}$ is a real symmetric matrix, $\mu_i \ge 0$. According to \underline{A4}, $\mu_{min} = 0$, thus $\zeta_{max}=1$ and $\zeta_{i} \le 1$. Besides, since $\zeta_1 \ge \zeta_2 \ge \cdots \ge \zeta_{m}$, we can observe that $\zeta_1 = 1$, and $\lambda_{\mathrm{max}}\left[(\mathbf{I} - \mathbf{J})\mathbf{P}^{2e}(\mathbf{I} - \mathbf{J})\right] = \zeta_2^{2e}$. Moreover, we can obtain 
		\begin{equation}
			\label{eq:result_algebraic_connectivity}
			\left\|\mathbf{P}^{e} (\mathbf{I} - \mathbf{J}) \right\|_{\mathrm{op}}^2 = \left[1-\epsilon \mu_{2}(\mathbf{La})\right]^{2e},
		\end{equation}
		where $\mu_{2}$ is the second smallest eigenvalue of $\mathbf{La}$. By substituting \eqref{eq:result_algebraic_connectivity} into \eqref{eq:result-gradient-norm_consensus}, we can further obtain
		\begin{align}
			 & \sum_{i=1}^{m} \mathbb{E} \left\| \bar{\theta}_k - \theta_{k}^{(i)} \right\|^2  \le \underbrace{\eta^2 \mathbb{E} \left\| Y_j \right\|_{\mathrm{F}}^2 }_{(a)}\ \left[1-\epsilon \mu_{2}(\mathbf{La})\right]^{2e}. \label{eq:result_theta_update_consensus}
		\end{align}
		We can observe that the term $(a)$ of \eqref{eq:result_theta_update_consensus} also emerges in \eqref{eq:expected_theta_error} in the proof of Appendix B. Therefore, we can directly get
		\begin{align}
			 & \frac{L^2}{mK} \sum_{k=0}^{K-1} \sum_{i=1}^{m} \mathbb{E} \left\| \bar{\theta}_k - \theta_{k}^{(i)} \right\|^2 \le \eta^2 \sigma^2 L^2 (\tau + 1) \left[1-\epsilon \mu_{2}(\mathbf{La})\right]^{2e} \notag \\
			 & +  \left[2\eta^2 L^2 \tau \beta + \eta^2 L^2 \tau (\tau + 1)\right] \left[1-\epsilon \mu_{2}(\mathbf{La})\right]^{2e} \notag                                                                               \\ &\frac{1}{mK} \sum_{k=0}^{K-1} \sum_{i=1}^{m} \left\| \nabla F(\theta_{k}^{(i)}) \right\|^2 \notag\\
			 & \le \eta^2 \sigma^2 L^2 (\tau + 1) \left[1-\epsilon \mu_{2}(\mathbf{La})\right]^{2e} \notag                                                                                                                \\
			 & +  \left[2\eta^2 L^2 \tau \beta + \eta^2 L^2 \tau (\tau + 1)\right] \frac{1}{mK} \sum_{k=0}^{K-1} \sum_{i=1}^{m} \left\| \nabla F(\theta_{k}^{(i)}) \right\|^2. \notag
		\end{align}
		By substituting the above equality into the term $(a)$ of \eqref{eq:basic-result-gradient-norm} in Lemma 4, we get the conclusion in Theorem 2.
	\end{proof}
\end{appendices}
\bibliographystyle{IEEEtran}
\bibliography{ref}

\begin{IEEEbiography}[{\includegraphics[width=1in,height=1.25in,clip,keepaspectratio]{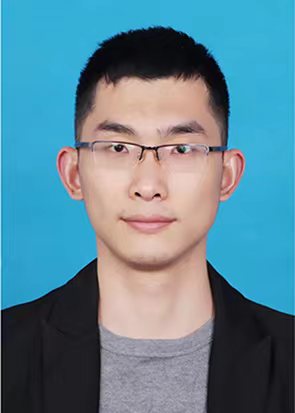}}]{Xing Xu}
	received the B.E. degree in Communication Engineering and the Ph.D. degree in Information and Communication Engineering from Huazhong University of Science and Technology and Zhejiang University, respectively. His research interests include collective intelligence, deep reinforcement learning, data analysis, and artificial intelligence.
\end{IEEEbiography}

\begin{IEEEbiography}[{\includegraphics[width=1in,height=1.25in,clip,keepaspectratio]{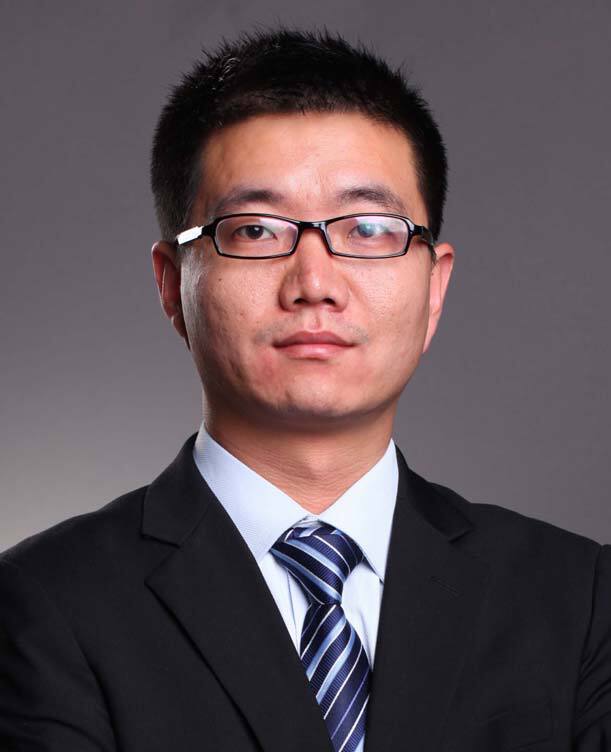}}]{Rongpeng Li}
	is currently an Associate Professor with the College of Information Science and Electronic Engineering, Zhejiang University, Hangzhou, China. He was a Research Engineer with the Wireless Communication Laboratory, Huawei Technologies Company, Ltd., Shanghai, China, from August 2015 to September 2016. He was a Visiting Scholar with the Department of Computer Science and Technology, University of Cambridge, Cambridge, U.K., from February 2020 to August 2020. His research interest currently focuses on networked intelligence for communications evolving (NICE). He received the Wu Wenjun Artificial Intelligence Excellent Youth Award in 2021. He serves as an Editor for \emph{China Communications}.
\end{IEEEbiography}

\begin{IEEEbiography}[{\includegraphics[width=1in,height=1.25in,clip,keepaspectratio]{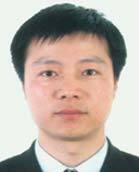}}]{Zhifeng Zhao}
	received the B.E. degree in computer science, the M.E. degree in communication and information systems, and the Ph.D. degree in communication and information systems from the PLA University of Science and Technology, Nanjing, China, in 1996, 1999, and 2002, respectively. From 2002 to 2004, he acted as a Post-Doctoral Researcher with Zhejiang University, Hangzhou, China, where his researches were focused on multimedia next-generation networks (NGNs) and softswitch technology for energy efficiency. From 2005 to 2006, he acted as a Senior Researcher with the PLA University of Science and Technology, where
	he performed research and development on advanced energy-efficient wireless router, \emph{ad-hoc} network simulator, and cognitive mesh networking test-bed.
	From 2006 to 2019, he was an Associate Professor with the College of Information Science and Electronic Engineering, Zhejiang University. Currently, he is with the Zhejiang Lab, Hangzhou as the Chief Engineering Officer. His research areas include software defined networks (SDNs), wireless network in 6G, computing networks, and collective intelligence. He is the Symposium Co-Chair of ChinaCom 2009 and 2010. He is the Technical Program Committee (TPC) Co-Chair of the 10th IEEE International Symposium on Communication and Information Technology (ISCIT 2010).
\end{IEEEbiography}

\begin{IEEEbiography}[{\includegraphics[width=1in,height=1.25in,clip,keepaspectratio]{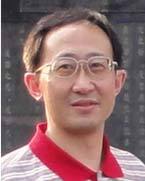}}]{Honggang Zhang} was an Honorary Visiting Professor with the University of York, York, U.K., and an International Chair Professor of excellence with the Université Européenne de Bretagne and Supélec, France. He is the Chief Managing Editor of Intelligent Computing, a Science Partner Journal, as well as a Professor with the College of Information Science and Electronic Engineering, Zhejiang University, Hangzhou, China. He has coauthored and edited two books: \emph{Cognitive Communications: Distributed Artificial Intelligence (DAI), Regulatory Policy \& Economics, Implementation} (John Wiley \& Sons) and \emph{Green Communications: Theoretical Fundamentals, Algorithms and Applications}  (CRC Press), respectively. His research interests include cognitive radio and networks, green communications, mobile computing, machine learning, artificial intelligence, and the Internet of Intelligence (IoI). He is a co-recipient of the 2021 IEEE Communications Society Outstanding Paper Award and the 2021 IEEE \textsc{Internet of Things Journal} (IoT-J) Best Paper Award. He was the leading Guest Editor for the Special Issues on Green Communications of the \emph{IEEE Communications Magazine}. He served as a Series Editor for the \emph{IEEE Communications Magazine} (Green Communications and Computing Networks Series) from 2015 to 2018 and the Chair of the Technical Committee on Cognitive Networks of the IEEE Communications Society from 2011 to 2012. He is the Associate Editor-in-Chief of \emph{China Communications}.
\end{IEEEbiography}
\end{document}